\documentclass[letterpaper]{article} % DO NOT CHANGE THIS
\usepackage{aaai25}  % DO NOT CHANGE THIS
\usepackage{times}  % DO NOT CHANGE THIS
\usepackage{helvet}  % DO NOT CHANGE THIS
\usepackage{courier}  % DO NOT CHANGE THIS
\usepackage[hyphens]{url}  % DO NOT CHANGE THIS
\usepackage{graphicx} % DO NOT CHANGE THIS
\usepackage{natbib}  % DO NOT CHANGE THIS AND DO NOT ADD ANY OPTIONS TO IT
\usepackage{caption} % DO NOT CHANGE THIS AND DO NOT ADD ANY OPTIONS TO IT
\usepackage{algorithm}
\usepackage{algorithmic}

\usepackage{newfloat}
\usepackage{listings}

\usepackage{graphicx}
\usepackage{booktabs}
\usepackage{tikz}
\usepackage{comment}
\usepackage{cite}
\usepackage{amsmath,amssymb} % define this before the line numbering.
\usepackage{color}
\usepackage{varwidth}

\usepackage{verbatim}
\usepackage{multirow}
\usepackage{booktabs}
\usepackage{adjustbox}
\usepackage{array}
\usepackage{mathtools}
\usepackage{enumitem}
\usepackage{kotex}
\usepackage{etoolbox} % for \newrobustcmd

\usepackage{newfloat}
\usepackage{listings}
\newrobustcmd{\etal}{\textit{et al.}}
\DeclareCaptionStyle{ruled}{labelfont=normalfont,labelsep=colon,strut=off} % DO NOT CHANGE THIS
\floatstyle{ruled}
\newfloat{listing}{tb}{lst}{}
\floatname{listing}{Listing}
\title{Enabling Region-Specific Control via Lassos in Point-Based Colorization}
\author{
    Sanghyeon Lee,\hspace{0.5cm}
    Jooyeol Yun,\hspace{0.5cm}
    Jaegul Choo
}
\affiliations{
    Korea Advanced Institute of Science and Technology (KAIST)\\
    Daejeon, Korea\\
    {\tt\small  \{shlee6825, blizzard072, jchoo\}@kaist.ac.kr}

}

\begin{document}

\maketitle
\begin{abstract}
Point-based interactive colorization techniques allow users to effortlessly colorize grayscale images using user-provided color hints. However, point-based methods often face challenges when different colors are given to semantically similar areas, leading to color intermingling and unsatisfactory results—an issue we refer to as color collapse. The fundamental cause of color collapse is the inadequacy of points for defining the boundaries for each color. To mitigate color collapse, we introduce a lasso tool that can control the scope of each color hint. Additionally, we design a framework that leverages the user-provided lassos to localize the attention masks. The experimental results show that using a single lasso is as effective as applying 4.18 individual color hints and can achieve the desired outcomes in 30\% less time than using points alone.
\end{abstract}

\section{Introduction}

Point-based interactive colorization~\cite{levin2004, side} on grayscale images aims to assist users in restoring colors by selecting and applying them to specific locations. 
The primary objective in training these models~\cite{zhang2017, unicolor} is to generate colorized images with minimal user interaction by effectively propagating the user-selected colors to relevant areas. 
For instance, a model can significantly reduce the user's effort by automatically colorizing an entire apple given a single hint. 
These models are not only useful for restoring aged photographs but also for a wide range of tasks, such as recoloring images and creating artistic visuals. 

\begin{figure}[t]
    \centering
    \includegraphics[width=\columnwidth]{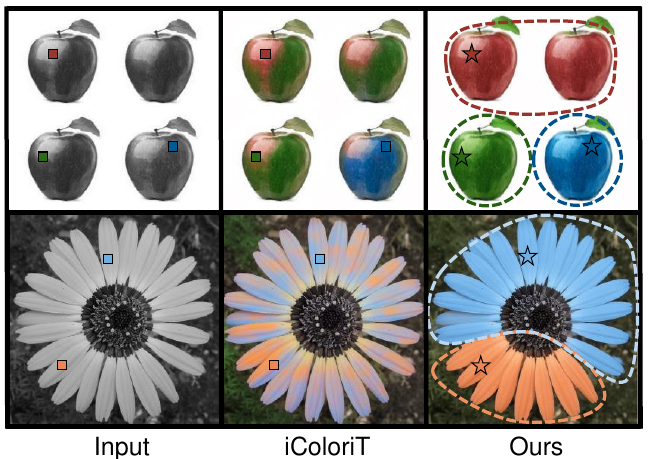}
    \caption{The examples of the color collapse. The start mark and the corresponding lasso in the same color describe the region designated for each color hint by the user. By specifying regions, users can better control how colors spread, thereby mitigating color collapse and leading to a more intentional colorization process.}
    \label{fig:main_intro_fail}
\end{figure}

However, existing methods often produce unsatisfactory images when multiple color hints are provided in closely related semantic regions. Specifically, Figure~\ref{fig:main_intro_fail} illustrates cases when different colors are assigned to semantically identical but separate objects (\textit{e.g.}, distinct apples or petals). As shown in the second column, even the state-of-the-art model~\cite{icolorit} suffers from the irregular intermingling of colors, producing implausible results. This issue, which we refer to as the \textbf{\textit{color collapse}}, arises as the model attempts to spread different colors across areas that appear similar. Color collapse is observed in repetitive patterns consisting of different colors ($\,e.g.$, flower petals, tiles, and fruit baskets). 

The fundamental cause of color collapse during point-based interactive colorization is the absence of interactive tools that enable the user to determine the region of the color hint spread. Although using color points as a medium of interaction is simple, these alone provide insufficient guidance for the model on the limits of color spread, often requiring excessive color hints to obtain a satisfactory result. Iteratively providing additional color hints until the color collapse is resolved is not only time-consuming but also diminishes user-friendliness.

Addressing this inherent problem with point-based interactions, we introduce an additional interactive tool, the lasso, which allows users to roughly define the scope of the color they want to spread. As shown in the third column of Figure~\ref{fig:main_intro_fail}, our lasso tool is designed to operate with loosely defined boundaries, eliminating the need for users to provide strict contours, which are often challenging to define. 

We first design a colorization model that uses cross-attention layers~\cite{attention} to inject color hints into the image. By restricting the cross-attention map to only attend within the user-provided lasso, we effectively control the scope of each color hint influence. This approach allows our model to adapt effectively to different lassos provided by users, even when using the same color hint, ensuring consistent results that align with varying user preferences.

To demonstrate the effectiveness of our interactive tool, we evaluate our approach in commonly encountered but challenging colorization scenarios. Our extensive experiments demonstrate that our model can effectively assist users in resolving color collapse using lassos while also maintaining the ability to produce colorful images without using lassos. Furthermore, our user study reveals that a single lasso interaction is as effective as 4.18 color points, and users achieve the same quality results in approximately \textit{30\% less time}. 

Our contributions are as follows:
\begin{itemize}
    \item We introduce a lasso tool that enhances point-interactive colorization by enabling users to precisely control the region where colors propagate, effectively addressing the color collapse. 
    \item We propose a framework incorporating a localization attention mask, effectively limiting the spread of color hints while adapting to various sizes of lassos.
    \item We demonstrate through experiments that our lasso tool reduces the number of interactions and time required for colorization tasks by effectively mitigating color collapse.
\end{itemize}

\section{Related Work}
\label{sec:reference}

\subsection{Interactive Colorization}
Interactive colorization models are designed to generate colorized images from grayscale input leveraging color conditions. Users can manipulate these conditions to tailor the colorized output to their preferences. Based on the precision with which the conditions are applied, these methods can be categorized into global and localized interactions. Global interactions alter the overall style of the image rather than focusing on specific locations. 

Widely studied global interactions include the use of example images where users select an image with a desired style to influence the global style of the colorization \cite{he2018deep, li2019automatic, zhang2019deep, xiao2020example, lu2020gray2colornet, xu2020stylization, li2021globally, yin2021yes, bai2022semantic}. Another approach involves modifying the color palette by adjusting the color histogram to apply a consistent color theme throughout the image \cite{wang2022palgan, wu2023flexicon}. Additionally, textual inputs allow users to specify color tones or themes globally using words that denote different colors \cite{chen2018language, bahng2018coloring, manjunatha2018learning, weng2022code, chang2022coder, chang2023coins}.

These global interactions are beneficial for their simplicity and minimal user effort, enabling changes in style with just a single global condition. However, they are designed primarily for global style modifications, thus limiting their capacity for detailed color editing on specific locations.

Conversely, localized interaction allows users to specify exact locations within an image to apply edits. A primary method for localized interaction is the use of points.
Traditional point-based colorization approaches~\cite{levin2004, side} employ hand-crafted image filters in an optimization-based approach. These methodologies require an optimization process tailored to each image, which prevents real-time modifications, significantly limiting their practicality.  
Learning-based methods have been developed to overcome the inefficiencies of inference time optimization. In early deep-learning-based research, Zhang~\etal~\cite{zhang2017} leverages a U-net structure to enable propagation based on semantic information. 
Recently, Yun~\etal~\cite{icolorit} achieved significant performance improvement by utilizing the long-range receptive fields of Vision Transformers~\cite{vit} to spread hint information to distant relevant areas. 
 
There are also efforts to integrate global and localized interactions to enhance the effectiveness of the colorization process ~\cite{unicolor, liang2024control}. 
However, despite these advancements, localized interaction models lack the ability to control the area over which a color hint spreads, often necessitating exhaustive trial-and-error until the desired coloration is achieved.

 \begin{figure*}[t]
    \centering
    \includegraphics[width=1\textwidth]{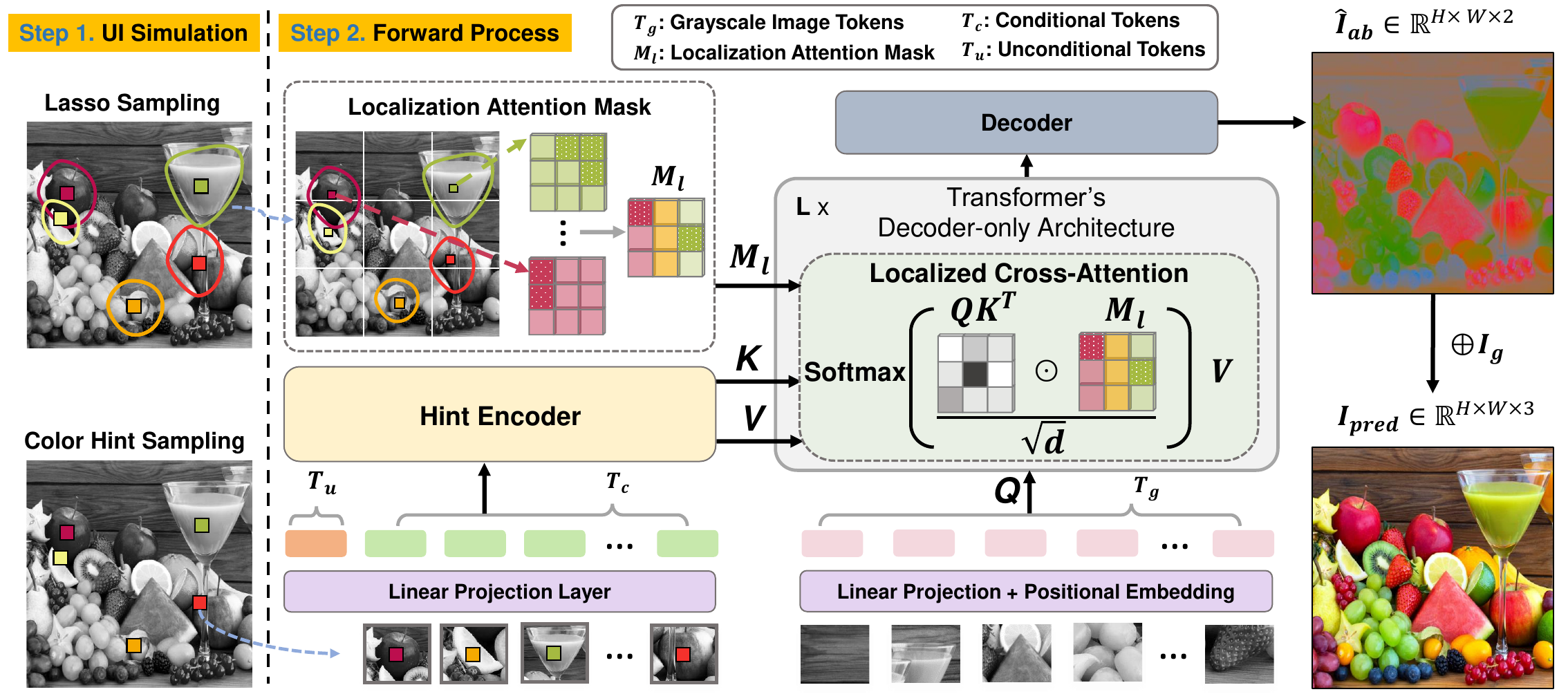}
    \caption{\textbf{The overview of our framework.} Our framework acquires color hints and corresponding lassos through a user interaction simulation process for training. The grayscale image is used as the query, and color hints as keys and values generate the cross-attention map $QK^T$. Subsequently, the attention map is modulated by an attention mask derived from the lassos to precisely control the influence of each color hint on the query image tokens.}
    \label{fig:method_main}
\end{figure*}

\subsection{Attention Manipulation}
Recent studies ~\cite{hertz2022prompt, xiao2023fastcomposer, park2022shape} have made advances in text-driven image editing by manipulating the attention maps of pre-trained large-scale text-to-image synthesis diffusion models~\cite{ldm}.
Hertz et al.~\cite{hertz2022prompt} propose a method that directly utilizes the attention maps derived from a reference image during the diffusion process, leading the generated image to faithfully capture the style of the reference image, for image editing by injecting attention maps, copied from a target condition, into the diffusion generation process. FastComposer~\cite{xiao2023fastcomposer} demonstrates similar findings in multi-subject personalization of diffusion model by fine-tuning a text-to-image generation model to create distinct attention map for individual subjects, resulting in successful generation across multiple subjects. Similarly, ~\cite{park2022shape} controls the shape of the generated image by directly masking within the attention map linked to the subject text token, influencing the spatial shape of the subject in the resultant image.

\section{Method}
\subsection{Overall Workflow}
 Figure~\ref{fig:method_main} illustrates the overall framework of our proposed methods, which utilize user-provided color hints and lasso input to colorize grayscale images. We leverage the L-channel from the CIELab image $I_{\text{Lab}}$ as the grayscale image to initiate the process. Our model incorporates a decoder structure inspired by Transformer~\cite{attention}. Grayscale images and user-provided color hints are transformed into patches and serve as the network’s inputs. We leverage the grayscale patches as the queries and the color hints as the keys and values in the cross-attention mechanism. This attention map shows which colors are propagated to which areas of the image. We use a lasso associated with each color hint to create an attention mask that controls the spread of colors based on the hints.

  In the final stage, the model predicts the ab color image $\hat{I}_{ab}$, which is then concatenated with the input grayscale image $I_g$ to produce the final output $I_{pred}$. Our framework employs a fixed-size pre-defined lasso to simplify the user’s task and improve usability. We determined the pre-defined lasso size by testing point increments and measuring PSNR on the benchmark dataset, selecting the size with the highest performance.
  This strategy allows users to refine the results through lassos only for color hints that do not accurately reflect their intentions.

\subsection{Simulating User Interactions During Training}
\label{subsec:simulation}
Our framework requires user-provided hints for training, but manually collecting extensive human data is infeasible. Instead, we simulate color hints and lassos from the ground truth image to mimic user behavior.

\noindent\textbf{Color hint simulation.} For simulating point interactions, we follow the sampling process from previous studies~\cite{zhang2017, icolorit}. Each color point consists of a hint location and ab color values, and the location is uniformly sampled from the image. During the training procedure, the number of hints $h$ is sampled from a uniform distribution $\mathcal{U}\sim(0,150)$.

\noindent\textbf{Lasso simulation.} Lassos determines the area affected by each point stroke. For each sampled color hint, we simulate a corresponding lasso. Specifically, the lasso is represented by an $H\times{W}$ binary mask, highlighting the regions needing attention. The intended scope of the hint can vary and be inaccurate due to individual differences. Thus, during training, we sample lassos from a randomly sized rectangle centered on the color hint’s location. With this approach, the ground truth color for the hint is always enclosed within the lasso region.

\begin{figure*}[t]
    \centering
    \includegraphics[width=1\textwidth]{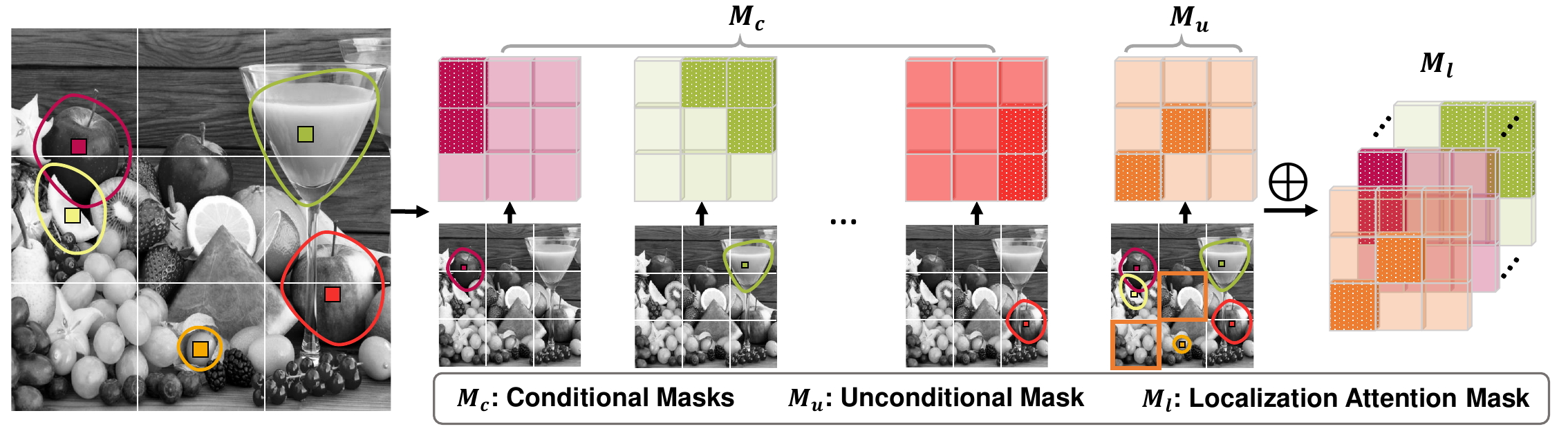}
    \caption{ \textbf{Localization Attention Mask.}   
      For each color hint, we apply a mask with a value of 1 to the tokens corresponding to patches interior of the lasso areas. Simultaneously, we construct an unconditional mask, $M_u$, based on regions not overlapped by lassos. The final localization attention mask, $M_l$, is produced by concatenating $M_u$ and $M_C$.}
    \label{fig:method_mask}
\end{figure*}

\subsection{Model Architecture}
Our model architecture consists of a hint encoder, localized cross-attention, and a decoder.

\noindent\textbf{Hint encoder.} The hint encoder accepts a specified number of $h$ color hints as input. To embed rich context information in each hint, we utilize 3-D cropped color hint patches of size $P\times{P}$, where $P$ denotes the patch size. To obtain the color hint patches $X_{hint}$, we crop patches centered on the color hints from the $I_{Lab}$. Within these cropped patches, areas not including the ab values at the points are masked with zeros.
These color hint patches are then embedded into conditional token $T_c$ through a linear projection layer. In our design, positional embedding is not directly applied to $X_{hint}$. Instead, the gray-scale patch $X_g$ provides positional information. The model can determine the precise locations for the color hints by leveraging the similarity of gray-scale between $X_g$ and $X_{hint}$. Beyond the conditional token $T_c$ produced by the linear projection layer, we incorporate an unconditional token $T_u$ into the hint encoder’s input. This unconditional token ensures the model’s operation when color hints are not provided. Containing information about the entire image, this token assists in coloring areas where no color hints are provided.

\noindent\textbf{Localized cross-attention.}
The localized cross-attention layer utilizes the structure of a transformer’s decoder-only architecture. In this layer, attention computation involves constructing query, key, and value representations. These are derived from grayscale image tokens, $T_g$, and hint tokens, $T_c$ and $T_u$. The grayscale image tokens $T_g$ are obtained by applying a linear projection layer and positional embeddings to the input grayscale image patches.

Specifically, the query matrix $Q$ is generated from these image tokens $T_g$, while the key and value matrices $K$ and $V$ are produced from the hint tokens. The dimensions of these matrices are defined as $Q\in\mathbb{R}^{N\times d}$ and $K, V\in\mathbb{R}^{(h+1)\times d}$, where $N$ is the number of image tokens, $h+1$ is the number of hint tokens, and $d$ is embedding dimension. 

The localized cross-attention operation is formulated as 
\begin{equation}
    \text{Attention}(Q,K,V) = \text{softmax}(QK^T / \sqrt{d} \odot M_l)V, 
\label{eq:lca}
\end{equation}
where $M_l$ is the localization attention mask from the sampled lassos.

\noindent\textbf{Localization attention mask.} 
To focus on the color-related region, as shown in Figure~\ref{fig:method_main}, localization attention mask $M_l$ explicitly masking the attention map $QK^T$ from the lassos. In the training process, these masks are driven from the simulated lassos. 

First, given the number of $h$ color hints, we resize each corresponding lasso $L\in \mathbb{R}^{H\times W\times h}$ into sizes with $H/P\times W/P$. Afterward, we define a conditional mask $M_c \in \mathbb{R}^{H/P \times W/P \times h}$ corresponding to the hint tokens $T_c$. As illustrated in Figure~\ref{fig:method_mask}, these conditional masks originate from the lasso, where the interior of the lasso is set to 1, while all other areas are set to 0. Meanwhile, the unconditional mask $M_u$ is a mask with the same spatial dimensions as $M_c$, designed to identify patches not specified by the lasso interaction. In this mask, areas not designated by the lasso are marked as 1, and all other areas are 0. Our final localization attention mask, $M_l \in \mathbb{N}^{(h+1) \times N}$, is constructed by concatenating $M_c$ and $M_h$, and then reshaped to match the size of the cross-attention map, where $N$ is the number of grayscale image tokens. 

\noindent\textbf{Decoder.}
The color image $\hat{I}_{ab}$ is then obtained using pixel shuffling~\cite{icolorit}, an efficient upsampling technique that rearranges the output feature map. Finally, $\hat{I}_{ab}$ is concatenated with the input grayscale image $I_{g}$, producing the predicted color image $I_{pred} \in \mathbb{R}^{H\times W\times 3}$.

\subsection{Objective Function}
 The lasso provides an attention mask that guides color propagation, ensuring the model colorizes only within the defined region and preventing undesirable color spread. Therefore, with no additional regularization term, we rely solely on the Huber loss~\cite{huber} between $I_{pred}$ and $I_{gt}$ in the CIELab color space. The Huber loss $L_h$, a conventional loss function within colorization tasks~\cite{cic, zhang2017, icolorit, ct2}, is computed as follows:
\begin{equation}
\begin{split}
    \mathcal{L}_{h} = &\frac{1}{2}(I_{pred} - I_{gt})^2 \textsc{I}_{\left|I_{pred} - I_{gt}\right| < 1} \\ 
    &+ (\left|I_{pred} - I_{gt}\right|-\frac{1}{2}) \textsc{I}_{\left|I_{pred} - I_{gt}\right| \geq 1}.
\end{split}
\end{equation}
\section{Experiments}

\begin{figure*}[t]
    \centering
    \includegraphics[width=1\linewidth]{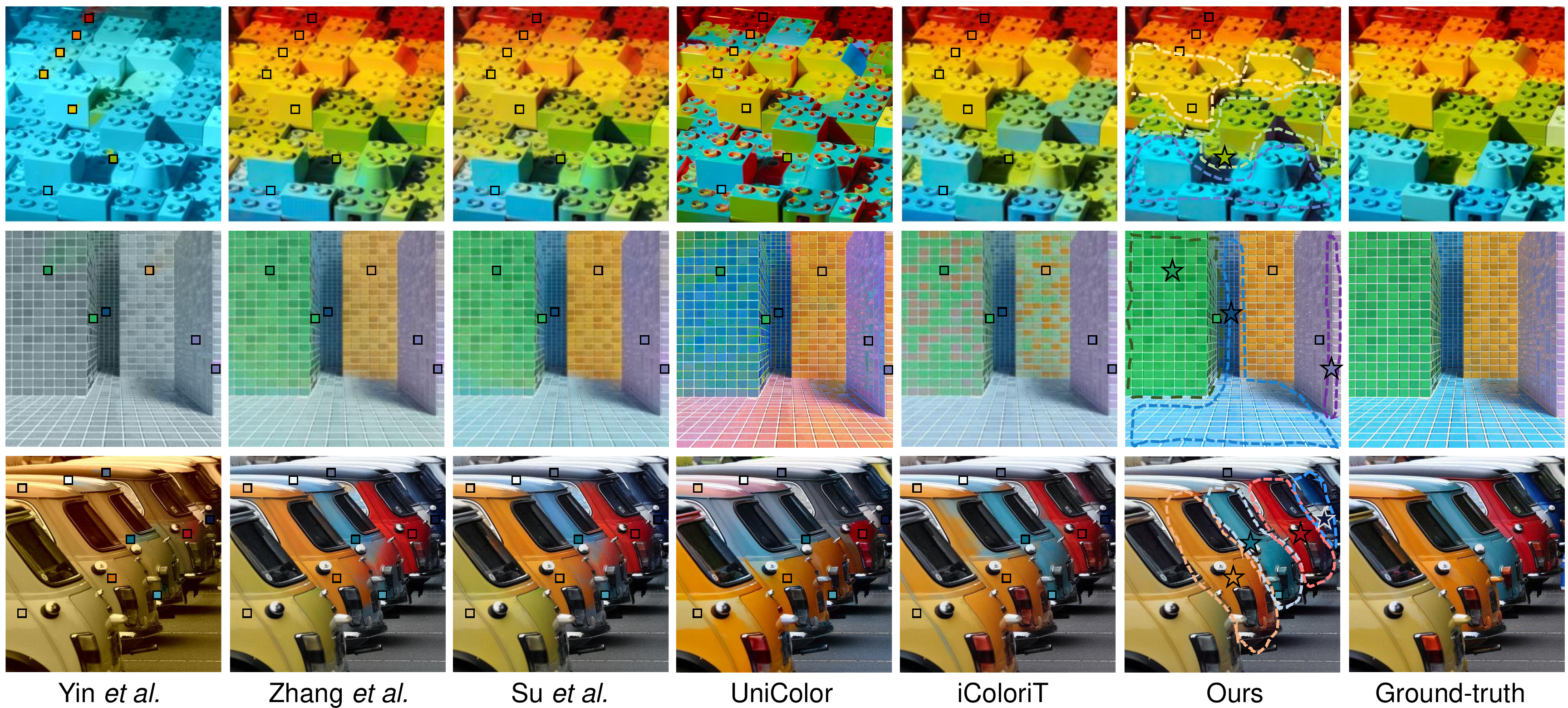}
    \caption{Qualitative results compare with baselines. Each star and its matching-colored lasso highlight the user-selected region for that color.  The presented results from our method reflect the colorization achieved through user-directed applications of both lassos and points with pre-defined lasso.}
    \label{fig:main_exp_real_quli}
\end{figure*}

\noindent\textbf{Datasets.} For the training process, we utilize the ImageNet 2012 dataset~\cite{imagenet}, which contains 1.3M images. We employ the ImageNet ctest~\cite{ctest} dataset, a commonly used benchmark in colorization research, to evaluate our approach. Also, we broaden our scope to diverse domains with the Oxford 102flowers~\cite{flowers} and CUB-200~\cite{cub} datasets. The ImageNet ctest~\cite{ctest} is a subset of the ImageNet validation set containing 10,000 images. The Oxford 102flowers dataset encompasses 1,020 images of flowers, and the CUB-200 dataset consists of 3,033 images representing 200 species of birds. 
Furthermore, to validate the effectiveness of our methods, we manually collect 98 samples from \url{unsplash.com} that exhibit repetitive patterns. We utilize the collected dataset for user studies. 

\begin{figure}[t]
    \centering
    \includegraphics[width=1\columnwidth]{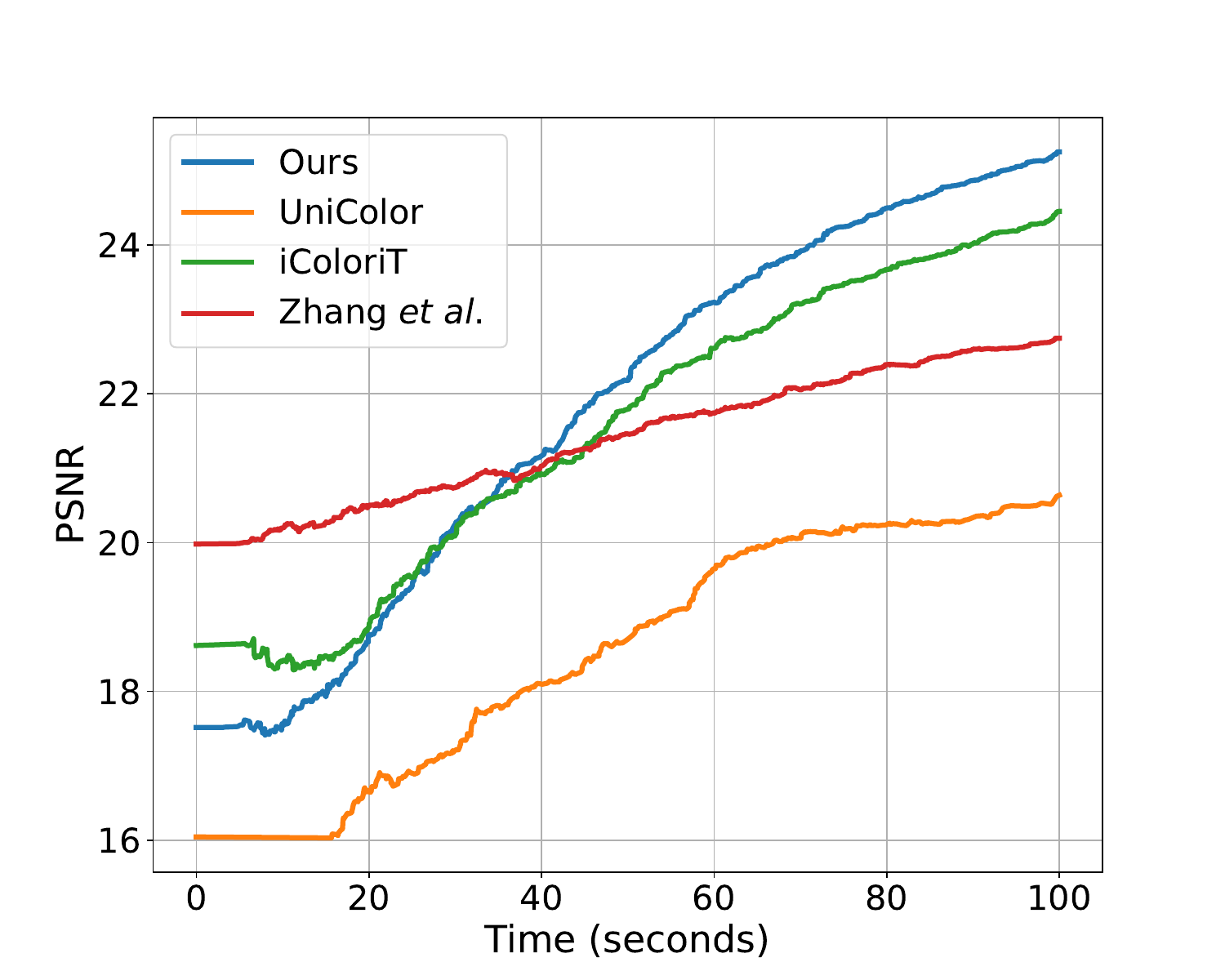}
    \caption{User study results on color collapse easy samples. We measure the average PSNR over the user interaction time, with the initial PSNR derived from each model’s unconditional inference.}
    \label{fig:main_exp_hard}
\end{figure}

\noindent \textbf{Baselines.} In our experiments, we compare our model with the point-interactive colorization approaches~\cite{side, zhang2017, icolorit}. For iColoriT~\cite{icolorit}, we utilize the base model trained for 100 epochs on ImageNet~\cite{imagenet}. Following the Kim~\etal~\cite{degas}, we also modify an unconditional model by Su~\etal~\cite{instanceaware} to handle user-provided point strokes. Additionally, we compare against UniColor~\cite{unicolor}, which relies on large (16×16 pixel) points to ensure a strong control signal. This inherently limits its fine-editing capabilities; therefore, we have included UniColor in user studies and selected qualitative comparisons.

\noindent \textbf{Evaluation metrics.} 
To evaluate the quality of the results, we employ the Peak Signal-to-Noise Ratio (PSNR), which measures the mean squared error between the ground truth and predicted images. 

\begin{figure}[t]
    \centering
    \includegraphics[width=1\columnwidth]{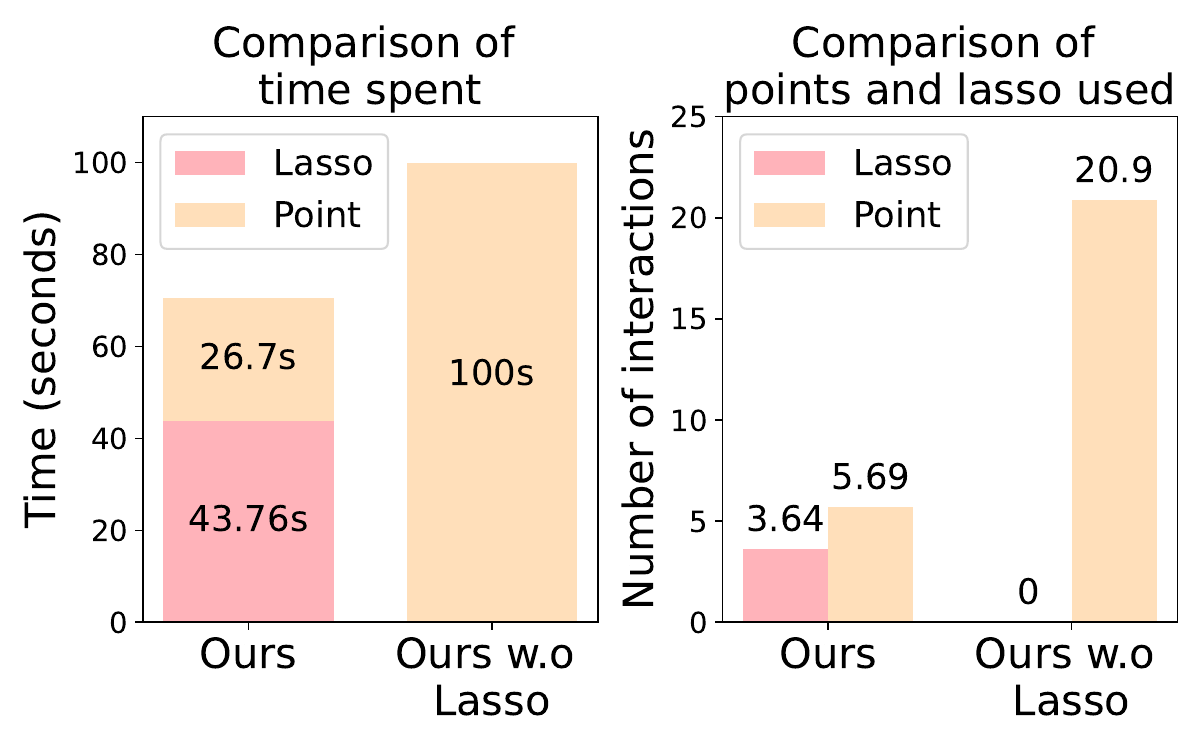}
    \caption{(Left) Time spent to achieve the same quality results with and without using lassos. (Right) The number of interactions required to achieve the same quality.}
    \label{fig:main_exp_wolasso}
\end{figure}

\begin{figure*}[h!]
    \centering
    \includegraphics[width=1\linewidth]{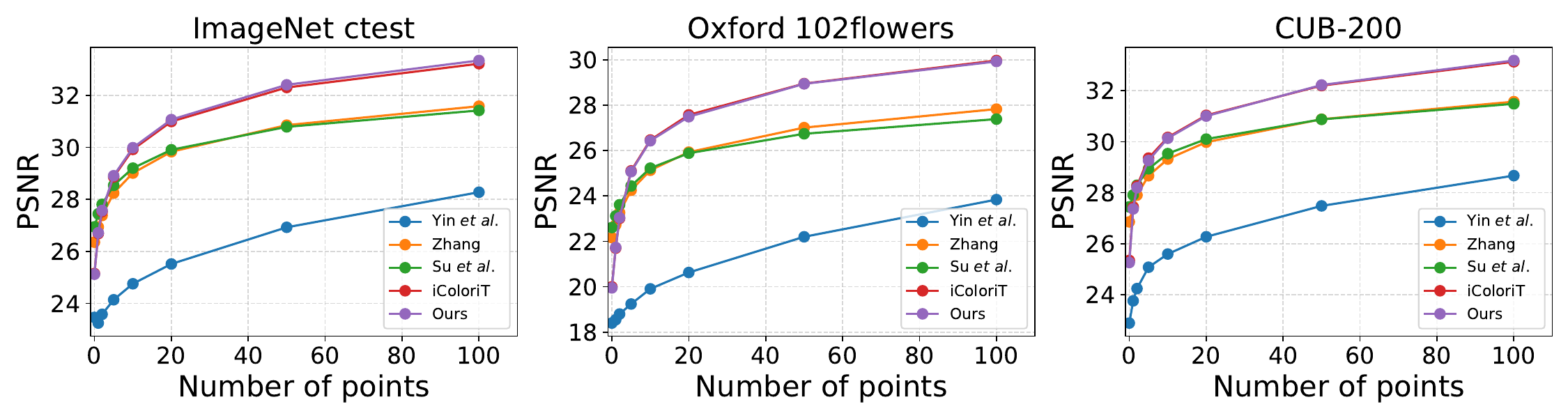}
    \caption{PSNR of the benchmark dataset according to the number of provided hints. Our method and the previous state-of-the-art iColoriT exhibit comparable performance using only point hints.}
    \label{fig:main_exp_psnr}
\end{figure*}

\begin{figure*}[t]
    \centering
    \includegraphics[width=1\linewidth]{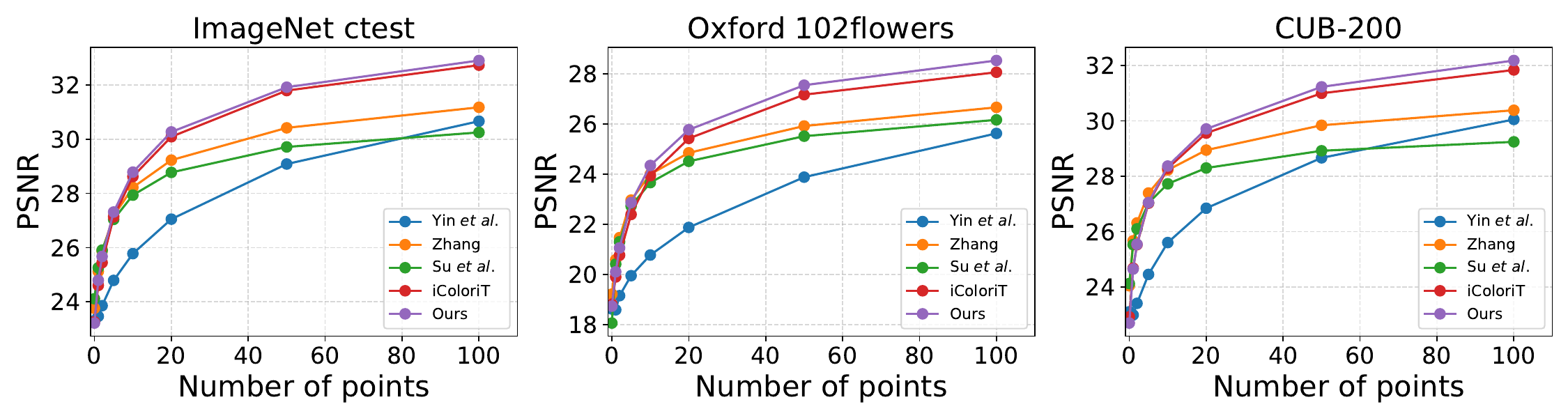}
    \caption{PSNR of the synthetic color collapse prone dataset according to the number of provided hints. A point pair refers to four points sampled from the same location across the $2\times 2$ grid images. } 
    \label{fig:main_exp_grid_psnr}
\end{figure*}

\subsection{Effectiveness Evaluation of the Lasso Tool}
\textbf{User study on handling color collapses.} 
We assess the efficacy of lasso interaction and investigate cases where it complements point-based interactions. For this study, we prepared a collection of 98 challenging samples characterized by similar patterns but varied colors and conducted a user study using this dataset. Human participants are provided with a user interface for colorization and asked to restore the grayscale image to its original color. Each user colorizes the image within a time limit of 100 seconds. If users did not provide a lasso interaction for the corresponding color hint, our model used a fixed-size, pre-defined lasso. To evaluate our approach, we conducted a comparison with baseline models that provided a user interface for colorization.

Figure~\ref{fig:main_exp_hard} presents the results of the user study, showing PSNR over time. The initial PSNR corresponds to the results of unconditional inference. In our models, the PSNR increases modestly as participants spend time drawing lassos. However, by the end of the 100-second period, our model with lasso interactions outperforms the baseline methods. In particular, UniColor~\cite{unicolor} achieves the lowest performance due to its longer inference times and difficulties with detailed editing. Although Zhang~\etal~\cite{zhang2017} demonstrates strong initial performance, its improvement plateaus as more color hints are introduced.

Furthermore, 85.7$\%$ of participants reported that the lasso interaction helped achieve better colorization results. Additionally, Figure~\ref{fig:main_exp_real_quli} shows the qualitative results for the challenging samples. Notably, color collapse is observed in baseline models that do not leverage lasso interactions, particularly in the first row’s green tile (bottom of the third column and inside the fourth column) and in the second row’s differently colored cars.

Figure~\ref{fig:main_exp_wolasso} shows how incorporating the lasso interaction improves user efficiency. We first conducted a user study in which participants were asked to colorize images using only points within 100 seconds, achieving a baseline PSNR of 23.94. We then measured how quickly and with how many interactions our method could reach the same PSNR when both points and lassos were available. 

As the figure illustrates, using lassos enabled our method to reach the target PSNR 29.50 seconds faster than the point-only approach. Moreover, while the point-only setup required 20.9 points on average, the lasso-included setup used only 5.69 points and 3.64 lassos. Since the time taken per interaction is similar for both points and lassos, 3.64 lassos effectively replaced 15.12 points. This corresponds to one lasso providing the same effectiveness as approximately 4.18 $(4.18=15.12/3.64)$ points, clearly demonstrating the efficiency gains offered by integrating the lasso interaction.

\begin{figure}[t]
    \centering
    \includegraphics[width=0.8\columnwidth]{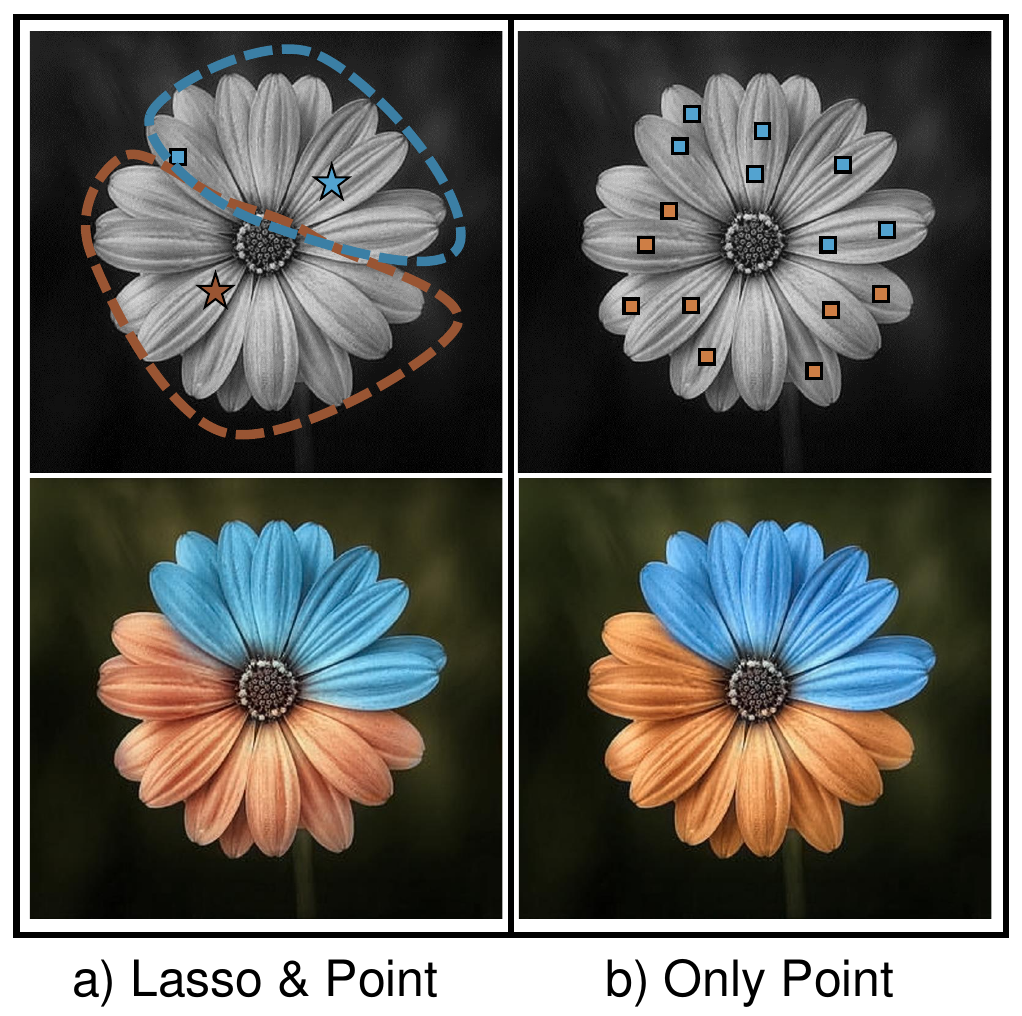}
    \caption{Comparison of colorization results using our model with lassos and points vs using only points.}
    \label{fig:main_exp_user_example}
\end{figure}

Figure ~\ref{fig:main_exp_user_example} presents an aesthetically appealing example generated using our model. To achieve a visually similar result, as judged by human perception, the process required three points and two lassos with corresponding points, and the user obtained the final output with only three inferences. In contrast, when using only color hints, as shown on the right, achieving the same result required 15 points, necessitating 15 iterations of model inference and correction.

\subsection{Comparison on Benchmark Datasets}
\noindent\textbf{Evaluation on colorization benchmarks.}
We conduct a systemic quantitative comparison against existing point interactive colorization models by randomly sampling points on test images and assigning their corresponding ground truth colors as hints. 
For our approach, we use a pre-defined lasso for each point to ensure fair evaluation conditions. As shown in Figure~\ref{fig:main_exp_psnr}, our model performs comparably to the previous state-of-the-art, even without the use of a lasso.

\noindent\textbf{Evaluation on the synthetic color collapse-prone dataset.} Color collapse is often encountered when distinct colors are provided to semantically similar objects within an image. 
To effectively simulate these scenarios, we synthesize a color collapse-prone dataset where the same pattern is repeated. As illustrated in the first row of Figure~\ref{fig:main_intro_fail}, we start by duplicating a single image to create four copies but with shifted colors. These are then organized into a $2\times 2$ grid. While each image keeps its original grayscale channel, we randomly change the color components (ab channel values) of each. 

To assess existing colorization approaches on this dataset, color hints sampled from the original image are uniformly applied to the corresponding locations in corresponding locations within the grid, assigning different color hints to semantically similar regions. We evaluated the model by sampling between 1 and 100 points from each grid. A point pair refers to four points sampled from the same location across the $2\times 2$ grid.

As shown in Figure~\ref{fig:main_exp_grid_psnr}, our proposed approach outperforms in these challenging scenarios. Further qualitative results can be found in the supplementary material. 

\begin{table}[h]
\centering
\resizebox{0.85\columnwidth}{!}{%z
\begin{tabular}{c|c|cc}
\toprule
 & Ours & w.o $M_l$& w.o $L$  \\ \midrule
PSNR@1  & \textbf{26.722} & 25.805 & 25.984 \\ 
PSNR@10 & \textbf{29.987} & 26.728 & 28.465 \\ 
PSNR@100 & \textbf{33.349} & 30.190 & 30.406 \\ \bottomrule
\end{tabular}%
}
\caption{Ablation study results. The size of the pre-defined lasso $L$ used in the ablation studies is $16\times16$.}
\label{table:ablation}
\end{table}
\noindent \textbf{Ablation studies.}
We conduct an ablation study on the effect of the localization attention mask $M_l$ during the training phase to demonstrate its beneficial impact on performance. We train the model without $M_l$ and then apply the localization attention mask during inference. 
The results, shown in the second column of Table~\ref{table:ablation}, indicate the model’s performance trained without $M_l$ but with lassos applied during inference. 
These results demonstrate that including a training phase yields better performance compared to training-free methods~\cite{hertz2022prompt, park2022shape} that modify the cross-attention map of the pre-trained model.
Meanwhile, the results displayed in the third column represent the inference of our model without the lasso, which is the same as using the image lasso. Severely inaccurate lasso sizes degrade the quality of the results.

\section{Limitations}
Our lasso interaction operates on the resolution of the attention maps, specifically $H/P \times W/P$, which limits the precision of lassos. This challenge can be effectively addressed by employing both lasso and point interactions in a complementary manner. In our framework, each point hint is processed by cropping a localized image patch centered on the point’s coordinate, allowing fine-grained distinctions even within the same resolution patch. By using the lasso to specify the coarse area and then refining details with points, our approach balances efficiency and precision, effectively mitigating color collapse and enhancing user control in the colorization process.

\section{Conclusion} 
In this study, we tackle the color collapse problem by integrating a lasso tool for point-based colorization. The lasso tool offers a practical solution to color collapse, an undesired behavior in point-based colorization models. Our framework currently employs a cross-attention mechanism to integrate user-provided color hints, and this architecture suggests a promising direction for future work, enabling the model to accommodate a broader range of user hints.

\section{Acknowledgements}
This work was supported by the Institute for Information \& communications Technology Promotion(IITP) grant funded by the Korean government(MSIT) (No.RS-2019-II190075 Artificial Intelligence Graduate School Program(KAIST)), and Artificial intelligence industrial convergence cluster development project funded by the Ministry of Science and ICT(MSIT, Korea) \& Gwangju Metropolitan City. 
\bibliography{aaai25}

\begin{thebibliography}{38}
\providecommand{\natexlab}[1]{#1}

\bibitem[{Bahng et~al.(2018)Bahng, Yoo, Cho, Park, Wu, Ma, and Choo}]{bahng2018coloring}
Bahng, H.; Yoo, S.; Cho, W.; Park, D.~K.; Wu, Z.; Ma, X.; and Choo, J. 2018.
\newblock Coloring with words: Guiding image colorization through text-based palette generation.
\newblock In \emph{Proceedings of the european conference on computer vision (eccv)}, 431--447.

\bibitem[{Bai et~al.(2022)Bai, Dong, Chai, Wang, Xu, and Yuan}]{bai2022semantic}
Bai, Y.; Dong, C.; Chai, Z.; Wang, A.; Xu, Z.; and Yuan, C. 2022.
\newblock Semantic-sparse colorization network for deep exemplar-based colorization.
\newblock In \emph{European Conference on Computer Vision}, 505--521. Springer.

\bibitem[{Chang et~al.(2022)Chang, Weng, Li, Li, and Shi}]{chang2022coder}
Chang, Z.; Weng, S.; Li, Y.; Li, S.; and Shi, B. 2022.
\newblock L-CoDer: Language-based colorization with color-object decoupling transformer.
\newblock In \emph{European Conference on Computer Vision}, 360--375. Springer.

\bibitem[{Chang et~al.(2023)Chang, Weng, Zhang, Li, Li, and Shi}]{chang2023coins}
Chang, Z.; Weng, S.; Zhang, P.; Li, Y.; Li, S.; and Shi, B. 2023.
\newblock L-CoIns: Language-Based Colorization With Instance Awareness.
\newblock In \emph{Proceedings of the IEEE/CVF Conference on Computer Vision and Pattern Recognition}, 19221--19230.

\bibitem[{Chen et~al.(2018)Chen, Shen, Gao, Liu, and Liu}]{chen2018language}
Chen, J.; Shen, Y.; Gao, J.; Liu, J.; and Liu, X. 2018.
\newblock Language-based image editing with recurrent attentive models.
\newblock In \emph{Proceedings of the IEEE Conference on Computer Vision and Pattern Recognition}, 8721--8729.

\bibitem[{Dosovitskiy et~al.(2020)Dosovitskiy, Beyer, Kolesnikov, Weissenborn, Zhai, Unterthiner, Dehghani, Minderer, Heigold, Gelly et~al.}]{vit}
Dosovitskiy, A.; Beyer, L.; Kolesnikov, A.; Weissenborn, D.; Zhai, X.; Unterthiner, T.; Dehghani, M.; Minderer, M.; Heigold, G.; Gelly, S.; et~al. 2020.
\newblock An Image is Worth 16x16 Words: Transformers for Image Recognition at Scale.
\newblock In \emph{ICLR}.

\bibitem[{He et~al.(2018)He, Chen, Liao, Sander, and Yuan}]{he2018deep}
He, M.; Chen, D.; Liao, J.; Sander, P.~V.; and Yuan, L. 2018.
\newblock Deep exemplar-based colorization.
\newblock \emph{ACM Transactions on Graphics (TOG)}, 37(4): 1--16.

\bibitem[{Hertz et~al.(2022)Hertz, Mokady, Tenenbaum, Aberman, Pritch, and Cohen-Or}]{hertz2022prompt}
Hertz, A.; Mokady, R.; Tenenbaum, J.; Aberman, K.; Pritch, Y.; and Cohen-Or, D. 2022.
\newblock Prompt-to-prompt image editing with cross attention control.
\newblock \emph{arXiv preprint arXiv:2208.01626}.

\bibitem[{Huang, Zhao, and Liao(2022)}]{unicolor}
Huang, Z.; Zhao, N.; and Liao, J. 2022.
\newblock Unicolor: A unified framework for multi-modal colorization with transformer.
\newblock \emph{ACM Transactions on Graphics (TOG)}, 41(6): 1--16.

\bibitem[{Huber(1992)}]{huber}
Huber, P.~J. 1992.
\newblock Robust estimation of a location parameter.
\newblock In \emph{Breakthroughs in statistics}, 492--518. Springer.

\bibitem[{Kim et~al.(2021)Kim, Lee, Park, Choi, Seo, and Choo}]{degas}
Kim, E.; Lee, S.; Park, J.; Choi, S.; Seo, C.; and Choo, J. 2021.
\newblock Deep Edge-Aware Interactive Colorization against Color-Bleeding Effects.
\newblock In \emph{Proceedings of the IEEE/CVF International Conference on Computer Vision}, 14667--14676.

\bibitem[{Larsson, Maire, and Shakhnarovich(2016)}]{ctest}
Larsson, G.; Maire, M.; and Shakhnarovich, G. 2016.
\newblock Learning Representations for Automatic Colorization.
\newblock In \emph{European Conference on Computer Vision (ECCV)}.

\bibitem[{Levin, Lischinski, and Weiss(2004)}]{levin2004}
Levin, A.; Lischinski, D.; and Weiss, Y. 2004.
\newblock Colorization Using Optimization.
\newblock \emph{ACM Transactions on Graphics}, 23: 689–694.

\bibitem[{Li et~al.(2019)Li, Lai, John, and Rosin}]{li2019automatic}
Li, B.; Lai, Y.-K.; John, M.; and Rosin, P.~L. 2019.
\newblock Automatic example-based image colorization using location-aware cross-scale matching.
\newblock \emph{IEEE Transactions on Image Processing}, 28(9): 4606--4619.

\bibitem[{Li et~al.(2021)Li, Sheng, Li, Ali, and Chen}]{li2021globally}
Li, H.; Sheng, B.; Li, P.; Ali, R.; and Chen, C.~P. 2021.
\newblock Globally and Locally Semantic Colorization via Exemplar-Based Broad-GAN.
\newblock \emph{IEEE Transactions on Image Processing}, 30: 8526--8539.

\bibitem[{Liang et~al.(2024)Liang, Li, Zhou, Li, and Loy}]{liang2024control}
Liang, Z.; Li, Z.; Zhou, S.; Li, C.; and Loy, C.~C. 2024.
\newblock Control Color: Multimodal Diffusion-based Interactive Image Colorization.
\newblock \emph{arXiv preprint arXiv:2402.10855}.

\bibitem[{Lu et~al.(2020)Lu, Yu, Peng, Zhao, and Wang}]{lu2020gray2colornet}
Lu, P.; Yu, J.; Peng, X.; Zhao, Z.; and Wang, X. 2020.
\newblock Gray2colornet: Transfer more colors from reference image.
\newblock In \emph{Proceedings of the 28th ACM International Conference on Multimedia}, 3210--3218.

\bibitem[{Manjunatha et~al.(2018)Manjunatha, Iyyer, Boyd-Graber, and Davis}]{manjunatha2018learning}
Manjunatha, V.; Iyyer, M.; Boyd-Graber, J.; and Davis, L. 2018.
\newblock Learning to color from language.
\newblock \emph{arXiv preprint arXiv:1804.06026}.

\bibitem[{Nilsback and Zisserman(2008)}]{flowers}
Nilsback, M.-E.; and Zisserman, A. 2008.
\newblock Automated Flower Classification over a Large Number of Classes.
\newblock In \emph{Indian Conference on Computer Vision, Graphics and Image Processing}.

\bibitem[{Park et~al.(2022)Park, Luo, Toste, Azadi, Liu, Karalashvili, Rohrbach, and Darrell}]{park2022shape}
Park, D.~H.; Luo, G.; Toste, C.; Azadi, S.; Liu, X.; Karalashvili, M.; Rohrbach, A.; and Darrell, T. 2022.
\newblock Shape-Guided Diffusion with Inside-Outside Attention.
\newblock \emph{arXiv preprint arXiv:2212.00210}.

\bibitem[{Rombach et~al.(2022)Rombach, Blattmann, Lorenz, Esser, and Ommer}]{ldm}
Rombach, R.; Blattmann, A.; Lorenz, D.; Esser, P.; and Ommer, B. 2022.
\newblock High-resolution image synthesis with latent diffusion models.
\newblock In \emph{Proceedings of the IEEE/CVF conference on computer vision and pattern recognition}, 10684--10695.

\bibitem[{Russakovsky et~al.(2015)Russakovsky, Deng, Su, Krause, Satheesh, Ma, Huang, Karpathy, Khosla, Bernstein, Berg, and Fei-Fei}]{imagenet}
Russakovsky, O.; Deng, J.; Su, H.; Krause, J.; Satheesh, S.; Ma, S.; Huang, Z.; Karpathy, A.; Khosla, A.; Bernstein, M.; Berg, A.~C.; and Fei-Fei, L. 2015.
\newblock {ImageNet Large Scale Visual Recognition Challenge}.
\newblock \emph{International Journal of Computer Vision (IJCV)}, 115(3): 211--252.

\bibitem[{Su, Chu, and Huang(2020)}]{instanceaware}
Su, J.-W.; Chu, H.-K.; and Huang, J.-B. 2020.
\newblock Instance-aware image colorization.
\newblock In \emph{Proceedings of the IEEE/CVF Conference on Computer Vision and Pattern Recognition}, 7968--7977.

\bibitem[{Vaswani et~al.(2017)Vaswani, Shazeer, Parmar, Uszkoreit, Jones, Gomez, Kaiser, and Polosukhin}]{attention}
Vaswani, A.; Shazeer, N.; Parmar, N.; Uszkoreit, J.; Jones, L.; Gomez, A.~N.; Kaiser, {\L}.; and Polosukhin, I. 2017.
\newblock Attention is all you need.
\newblock \emph{Advances in neural information processing systems}, 30.

\bibitem[{Wang et~al.(2022)Wang, Xia, Qi, Shao, and Qiao}]{wang2022palgan}
Wang, Y.; Xia, M.; Qi, L.; Shao, J.; and Qiao, Y. 2022.
\newblock PalGAN: Image colorization with palette generative adversarial networks.
\newblock In \emph{European Conference on Computer Vision}, 271--288. Springer.

\bibitem[{Welinder et~al.(2010)Welinder, Branson, Mita, Wah, Schroff, Belongie, and Perona}]{cub}
Welinder, P.; Branson, S.; Mita, T.; Wah, C.; Schroff, F.; Belongie, S.; and Perona, P. 2010.
\newblock Caltech-UCSD Birds 200.
\newblock Technical Report CNS-TR-2010-001, California Institute of Technology.

\bibitem[{Weng et~al.(2022{\natexlab{a}})Weng, Sun, Li, Li, and Shi}]{ct2}
Weng, S.; Sun, J.; Li, Y.; Li, S.; and Shi, B. 2022{\natexlab{a}}.
\newblock CT 2: Colorization transformer via color tokens.
\newblock In \emph{European Conference on Computer Vision}, 1--16. Springer.

\bibitem[{Weng et~al.(2022{\natexlab{b}})Weng, Wu, Chang, Tang, Li, and Shi}]{weng2022code}
Weng, S.; Wu, H.; Chang, Z.; Tang, J.; Li, S.; and Shi, B. 2022{\natexlab{b}}.
\newblock L-code: Language-based colorization using color-object decoupled conditions.
\newblock In \emph{Proceedings of the AAAI Conference on Artificial Intelligence}, volume~36, 2677--2684.

\bibitem[{Wu et~al.(2023)Wu, Yang, Xu, Liu, Yan, and Zhang}]{wu2023flexicon}
Wu, S.; Yang, Y.; Xu, S.; Liu, W.; Yan, X.; and Zhang, S. 2023.
\newblock FlexIcon: Flexible Icon Colorization via Guided Images and Palettes.
\newblock In \emph{Proceedings of the 31st ACM International Conference on Multimedia}, 8662--8673.

\bibitem[{Xiao et~al.(2020)Xiao, Han, Zhang, Qin, Wong, Han, and He}]{xiao2020example}
Xiao, C.; Han, C.; Zhang, Z.; Qin, J.; Wong, T.-T.; Han, G.; and He, S. 2020.
\newblock Example-Based Colourization Via Dense Encoding Pyramids.
\newblock In \emph{Computer Graphics Forum}, volume~39, 20--33. Wiley Online Library.

\bibitem[{Xiao et~al.(2024)Xiao, Yin, Freeman, Durand, and Han}]{xiao2023fastcomposer}
Xiao, G.; Yin, T.; Freeman, W.~T.; Durand, F.; and Han, S. 2024.
\newblock Fastcomposer: Tuning-free multi-subject image generation with localized attention.
\newblock \emph{International Journal of Computer Vision}, 1--20.

\bibitem[{Xu et~al.(2020)Xu, Wang, Fang, Sheng, and Zhang}]{xu2020stylization}
Xu, Z.; Wang, T.; Fang, F.; Sheng, Y.; and Zhang, G. 2020.
\newblock Stylization-based architecture for fast deep exemplar colorization.
\newblock In \emph{Proceedings of the IEEE/CVF conference on computer vision and pattern recognition}, 9363--9372.

\bibitem[{Yin, Gong, and Qiu(2019)}]{side}
Yin, H.; Gong, Y.; and Qiu, G. 2019.
\newblock Side window filtering.
\newblock In \emph{Proceedings of the IEEE/CVF Conference on Computer Vision and Pattern Recognition}, 8758--8766.

\bibitem[{Yin et~al.(2021)Yin, Lu, Zhao, and Peng}]{yin2021yes}
Yin, W.; Lu, P.; Zhao, Z.; and Peng, X. 2021.
\newblock Yes," Attention Is All You Need", for Exemplar based Colorization.
\newblock In \emph{Proceedings of the 29th ACM International Conference on Multimedia}, 2243--2251.

\bibitem[{Yun et~al.(2023)Yun, Lee, Park, and Choo}]{icolorit}
Yun, J.; Lee, S.; Park, M.; and Choo, J. 2023.
\newblock iColoriT: Towards Propagating Local Hints to the Right Region in Interactive Colorization by Leveraging Vision Transformer.
\newblock In \emph{Proceedings of the IEEE/CVF Winter Conference on Applications of Computer Vision}, 1787--1796.

\bibitem[{Zhang et~al.(2019)Zhang, He, Liao, Sander, Yuan, Bermak, and Chen}]{zhang2019deep}
Zhang, B.; He, M.; Liao, J.; Sander, P.~V.; Yuan, L.; Bermak, A.; and Chen, D. 2019.
\newblock Deep exemplar-based video colorization.
\newblock In \emph{Proceedings of the IEEE/CVF conference on computer vision and pattern recognition}, 8052--8061.

\bibitem[{Zhang, Isola, and Efros(2016)}]{cic}
Zhang, R.; Isola, P.; and Efros, A.~A. 2016.
\newblock Colorful image colorization.
\newblock In \emph{European conference on computer vision}, 649--666. Springer.

\bibitem[{Zhang et~al.(2017)Zhang, Zhu, Isola, Geng, Lin, Yu, and Efros}]{zhang2017}
Zhang, R.; Zhu, J.-Y.; Isola, P.; Geng, X.; Lin, A.~S.; Yu, T.; and Efros, A.~A. 2017.
\newblock Real-time user-guided image colorization with learned deep priors.
\newblock \emph{ACM TOG}.

\end{thebibliography}

\clearpage
\section{Overview of Supplementary Materials}

This supplementary section extends the main manuscript by detailing further experiments that illustrate the effectiveness of various lasso sizes and presenting diverse qualitative results from our methods. Specifically, it includes:

\begin{itemize}
    \item \textbf{Implementation details:} Comprehensive descriptions of the network architecture, training procedures, parameter settings, and lasso size configurations used in our interactive colorization framework.
    
    \item \textbf{Effectiveness evaluation of the lasso:} Detailed analysis of how different lasso sizes impact performance.
    
    \item \textbf{Step-by-step demonstrations:} Illustrative examples showcasing how the integration of point and lasso interactions enhances precision in image editing.
    
    \item \textbf{Additional experimental results:} Expanded qualitative and quantitative evaluations demonstrating the strengths of our approach in diverse scenarios, including comparisons with baseline models, performance on synthetic color-collapse datasets, and examples of personalized colorization preferences.
    
\end{itemize}

\section{Implementation Details}

% \noindent\textbf{Implementation details.} 
In our study, we leverage the Transformer decoder architecture~\cite{attention} for the main network. The embedding dimension is 768, and the network depth is 12 layers. The architecture is designed to alternate between localized cross-attention and self-attention layers. The hint encoder employs a 12-layer MLP with a hidden dimension of 768. For image decoding, we use pixel shuffling from Yun~\etal~\cite{icolorit}. Specifically, the input images are resized to $224\times224$ and divided into patches of size $P=16$, resulting in an image token length of $N=HW/P^2$, which equals 196.

During the training, we utilize a variable number of color hints for each image, determined through random sampling from a uniform distribution $\mathcal{U}\sim(0,150)$. The placement of these color hints is uniformly distributed across the image. For batch processing, since the number of sampled points varies per image, we additionally pad the sequence with the maximum point number. These are then masked to prevent their entry into the attention map.
The size of each lasso, both height and width, are randomly sampled from a uniform distribution $\mathcal{U}(4, 64)$.

\section{Effectiveness of the Lasso Tool}
\noindent \textbf{Pre-defined lasso for point-interactive colorization.}
~\label{sec:pre-defined}
If the user does not make additional adjustments to the lasso interaction, our model functions in a point-interactive colorization manner by leveraging the fixed size of the lasso. This predetermined lasso size is $(P\times r)^2$, where $P$ is the size of the patch and $r$ is the scaling factor of the lasso area. We determined the optimal $r$ by testing point increments and measuring PSNR on the benchmark dataset, selecting the size that yields the highest performance. As shown in Table~\ref{table:supply_pre_defined}, larger lasso sizes perform better when the number of hints is small, while a smaller lasso size is more effective as the number of hints increases. This observation aligns with the understanding that higher hint density requires proportionally reduced areas for color spreading.

\begin{table}[h]
\centering
\resizebox{\columnwidth}{!}{%z
\begin{tabular}{ccccccc}
\toprule
 & r:0.25 & r:0.5 & r:1 & r:4 & r:16 & r:64 \\ \midrule
PSNR@1  & 26.716 & 26.720 & 26.722 & \textbf{26.723} &  26.317 & 26.003 \\ 
PSNR@10 & 29.965 & 29.975 & \textbf{29.987} & 29.961 & 28.802 & 28.477 \\ 
PSNR@100 & 33.326 & \textbf{33.353} & 33.349 & 33.132 & 30.887 & 30.417 \\ \bottomrule
\end{tabular}%
}
\caption{Quantitative results with various lasso sizes on ImageNet ctest. PSNR@K represents PSNR obtained with K number of color hints.}
\label{table:supply_pre_defined}
\end{table}

 \begin{figure}[h!]
    \centering
    \includegraphics[width=\columnwidth]{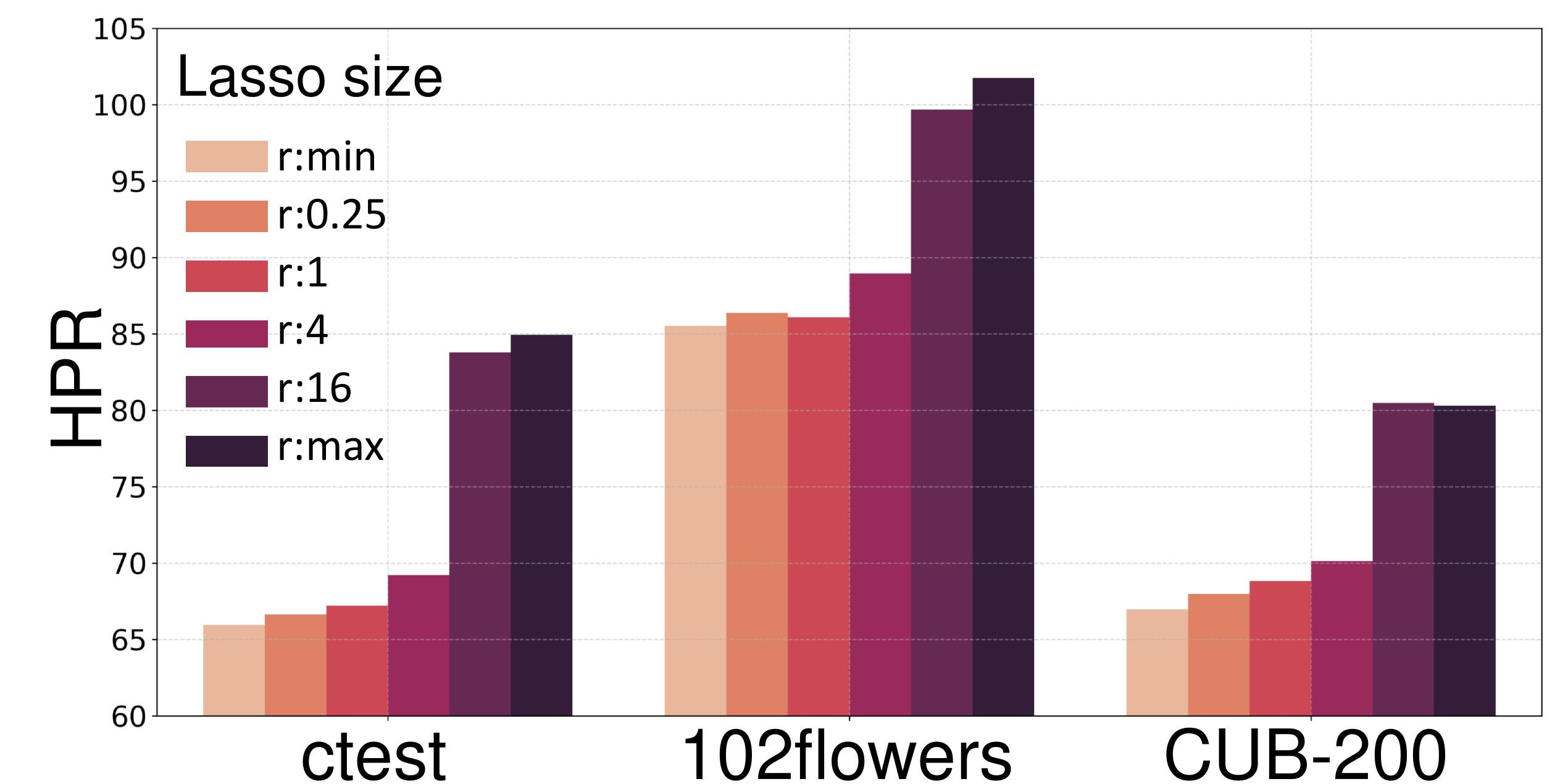}
    \caption{Average HPR on various lasso sizes. We calculated the average Hint HPR across three datasets. }
    \label{fig:supply_exp_hpr}
\end{figure}

\begin{figure}[t]
    \centering
    \includegraphics[width=0.9\columnwidth]{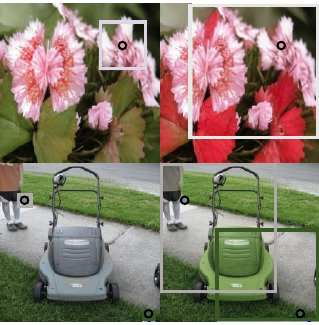}
    \caption{Qualitative results with various lasso sizes. With consistent color hints provided, users can adjust the range of color propagation from various lassos.}
    \label{fig:supply_exp_hpr_quli}
\end{figure}

\begin{figure*}[h!]
    \centering
    \includegraphics[width=\linewidth]{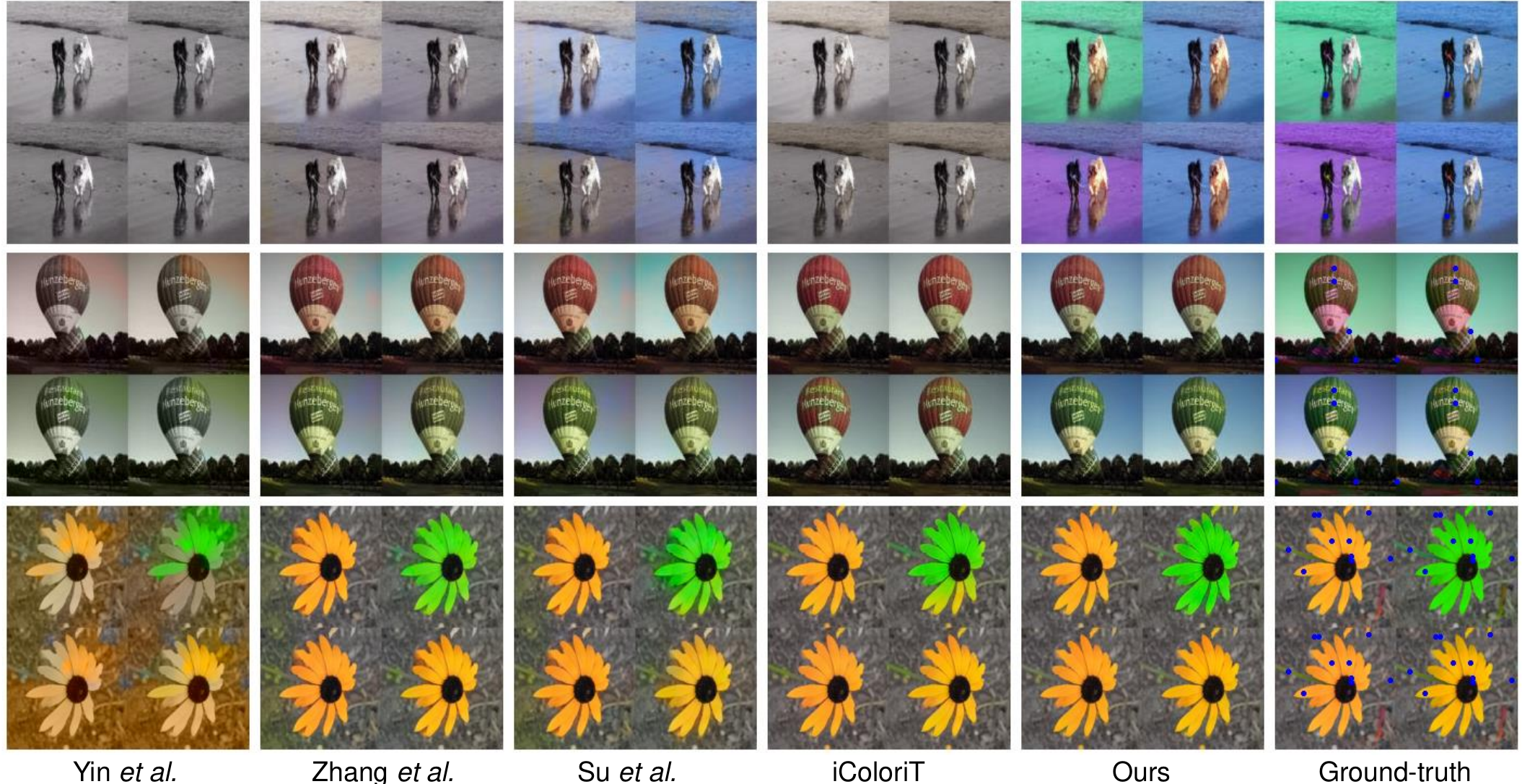}
    \caption{Qualitative results on the synthetic color-collapse dataset are presented with 1, 5, and 10 user-provided hints for each divided cell. For better visibility, the locations of the points on the ground truth are marked with blue dots.}
    \label{fig:supply_exp_grid_quali}
\end{figure*}

\noindent \textbf{Controlling the boldness of point strokes via lasso sizes.} 
Hint Propagation Range (HPR)~\cite{icolorit} is adept at identifying areas in a colored image that undergo cognitively noticeable changes from the newly provided hints. In scenarios where an image is edited by sequentially adding points, HPR evaluates the distance over which a color hint extends by comparing the image colored with the $t$-th hint to the image with the newly added point. This metric effectively demonstrates how the scope of a hint varies with different lasso sizes. 
Figure~\ref{fig:supply_exp_hpr} presents quantitative results for various lasso sizes. When using the minimum lasso size, attention masking is applied to just a single patch token. Conversely, when using the maximum lasso size, all areas are uniformly masked with a value of 1. 
 The propagation of color varies in accordance with the size of the lasso of the color hints. 
The colorized results in Figure~\ref{fig:supply_exp_hpr_quli} demonstrate the effects of applying different sizes of lassos on the same color hint. Even with imprecise lassos, selective color propagation within areas shares the same semantics.

\section{Step-by-Step Demonstration}
\begin{figure}[ht!]
    \centering
    \includegraphics[width=1\linewidth]{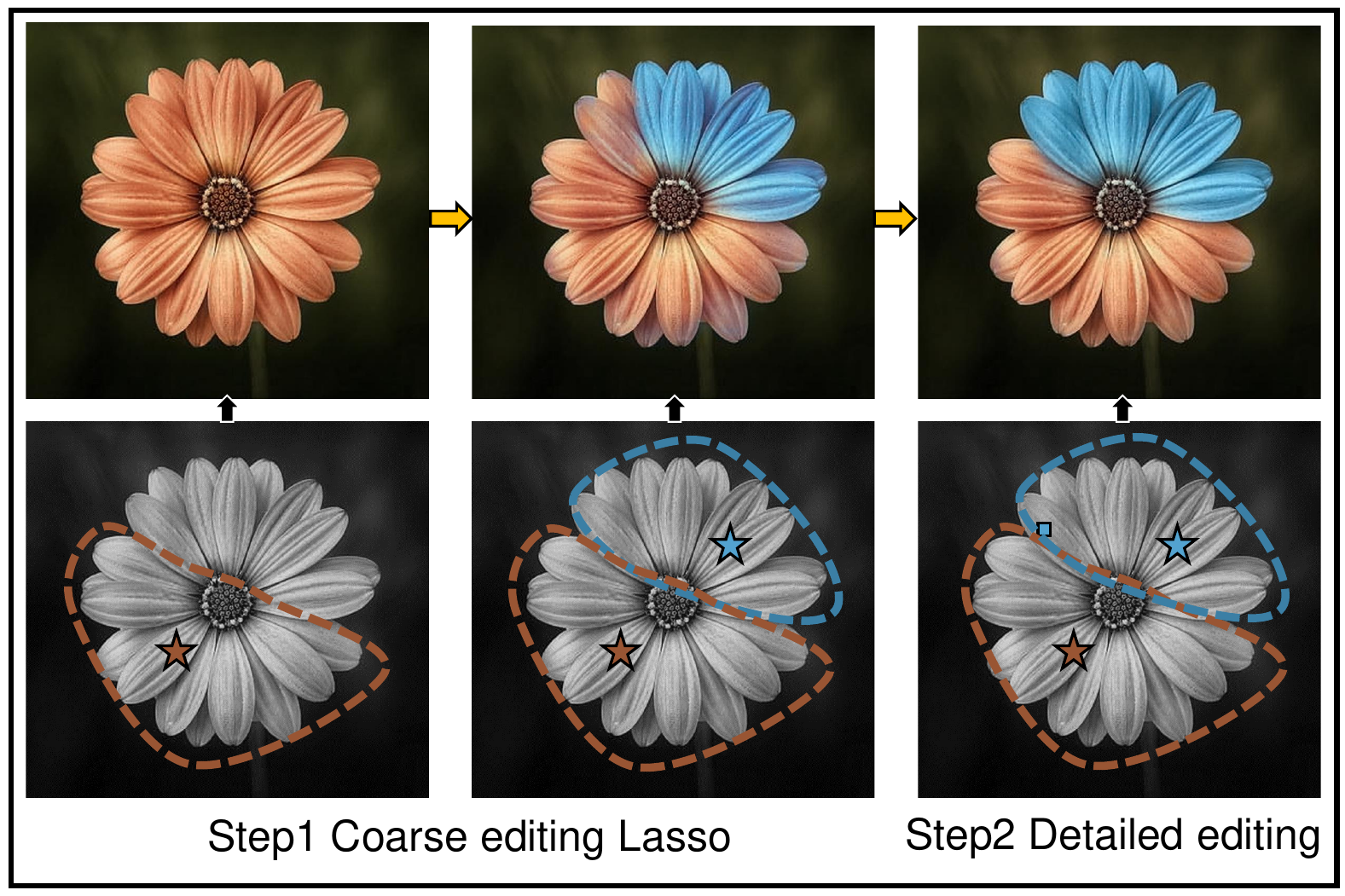}
    \caption{Step-by-step demonstration of enhanced precision using point and lasso interactions in image editing. Users begin by roughly defining areas and editing the detail region.}
    \label{fig:supply_coarse_to_fine}
\end{figure}

% \noindent \textbf{Step-by-step demonstration of our framework}
This section demonstrates how combining point and lasso interactions improves precision in image editing. As illustrated in Figure~\ref{fig:supply_coarse_to_fine}, the lasso interaction effectively propagates color across broader, coarser image regions. However, it may not accurately capture finer details within these areas. To overcome this limitation, point-based detailed editing is employed, allowing users to apply precise color changes to specific locations. This integrated approach reduces user effort by minimizing the number of interactions required and enhances the accuracy and quality of the final colorized output.

\section{Additional Experiments Results}
\noindent \textbf{Qualitative results of synthetic color collapse dataset.} Figure~\ref{fig:supply_exp_grid_quali} presents the outcomes on the synthetic two-by-two grid dataset with varying numbers of user-provided hints: 1, 5, and 10 for each divided cell. In the first row, baseline models either disregard the color hint (Yin~\etal~\cite{yin2021yes}, iColorit~\cite{icolorit}) or replicate the dominant color and ignore other colors (Zhang~\etal~\cite{zhang2017}, Su~\etal~\cite{instanceaware}). Conversely, Our model effectively propagates color hints to the correct region by employing a pre-defined lasso.

\begin{figure*}[!]
    \centering
    \includegraphics[width=1\linewidth]{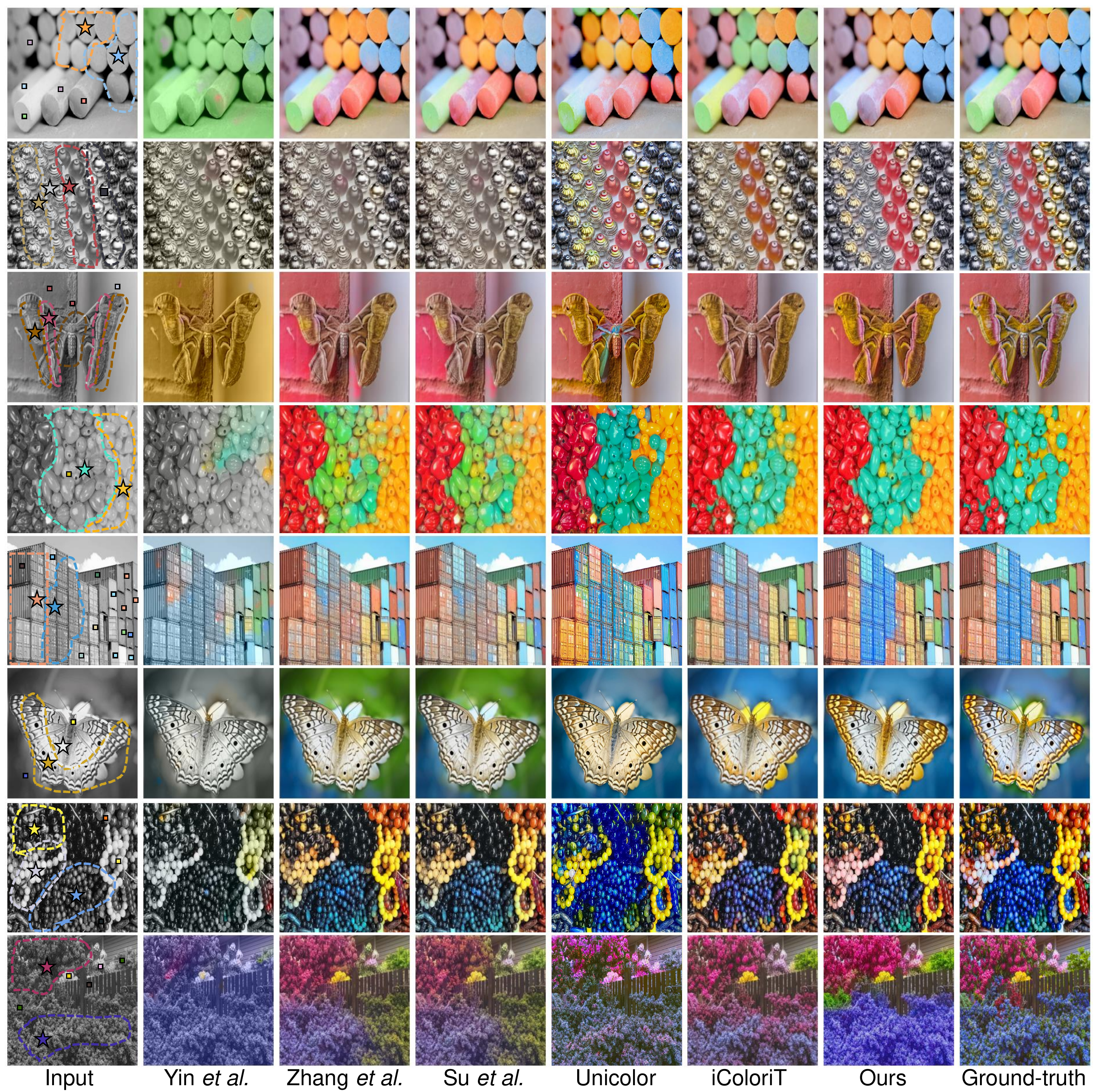}
    \caption{Qualitative results on the collected dataset. Baselines utilize only point hints, including those marked with a star for inference.}
    \label{fig:supply_real_world}
\end{figure*}

\begin{figure*}[t!]
    \centering
    \includegraphics[width=1\linewidth]{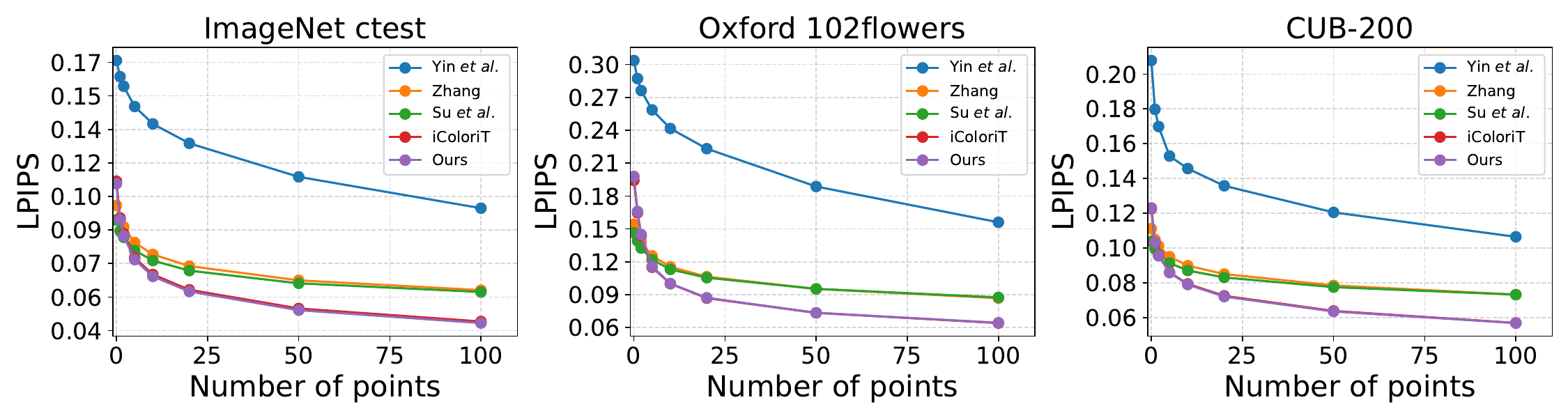}
    \caption{Quantitative results of point-interactive colorization methods. For a fair comparison, we utilize predefined localization, conducting inference based solely on the color hints without incorporating any extra user-defined localization.}
    \label{fig:supply_exp_lpips_aaai}
\end{figure*}

\begin{figure*}[t!]
    \centering
    \includegraphics[width=1\linewidth]{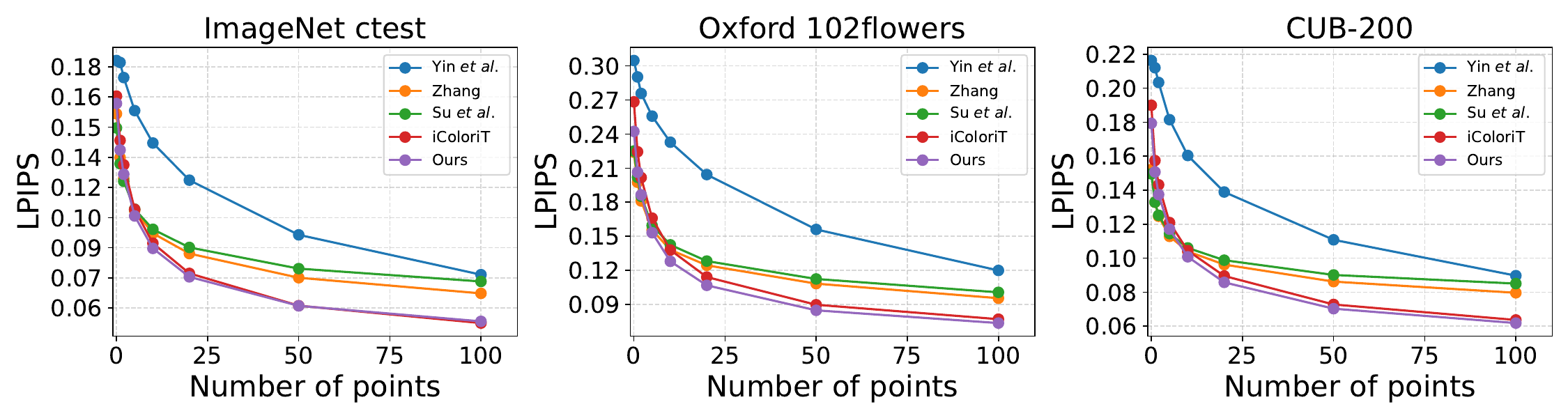}
    \caption{Quantitative results of point-interactive colorization methods on the 2 × 2 grid images. For a fair comparison, we utilize predefined localization, conducting inference based solely on the color hints without incorporating any extra user-defined localization.}
    \label{fig:supply_exp_lpips_grid_aaai}
\end{figure*}

\noindent \textbf{Additional quantitative results}. 
To assess the semantic similarity of the produced images, we measure the Learned Perceptual Image Patch Similarity (LPIPS)~\cite{lpips}. Figure~\ref{fig:supply_exp_lpips_aaai} presents the LPIPS scores for point-interactive colorization on a standard benchmark dataset, and 
Figure~\ref{fig:supply_exp_lpips_grid_aaai} illustrates the LPIPS scores for a synthetic dataset specifically designed to be prone to color collapse. These results demonstrate that our method achieves performance comparable to the existing state-of-the-art model, iColoriT~\cite{icolorit}, similar to our PSNR results.

\noindent \textbf{Comparison of the collected dataset.}
In Figure~\ref{fig:supply_real_world}, we compare the performance of our method on a dataset we collected, as detailed in the experiments section of the main manuscript. This comparison contrasts the outcomes of various baseline methods using user-provided color hints and lassos.

Notably, the baseline methods rely solely on color hints without incorporating lassos, with these hints directly derived from the ground truth. This comparison underscores the effectiveness of our lasso-integrated approach in enhancing colorization results by providing additional control and preventing color collapse.

\begin{figure*}[t!]
    \centering
    \includegraphics[width=0.9\linewidth]{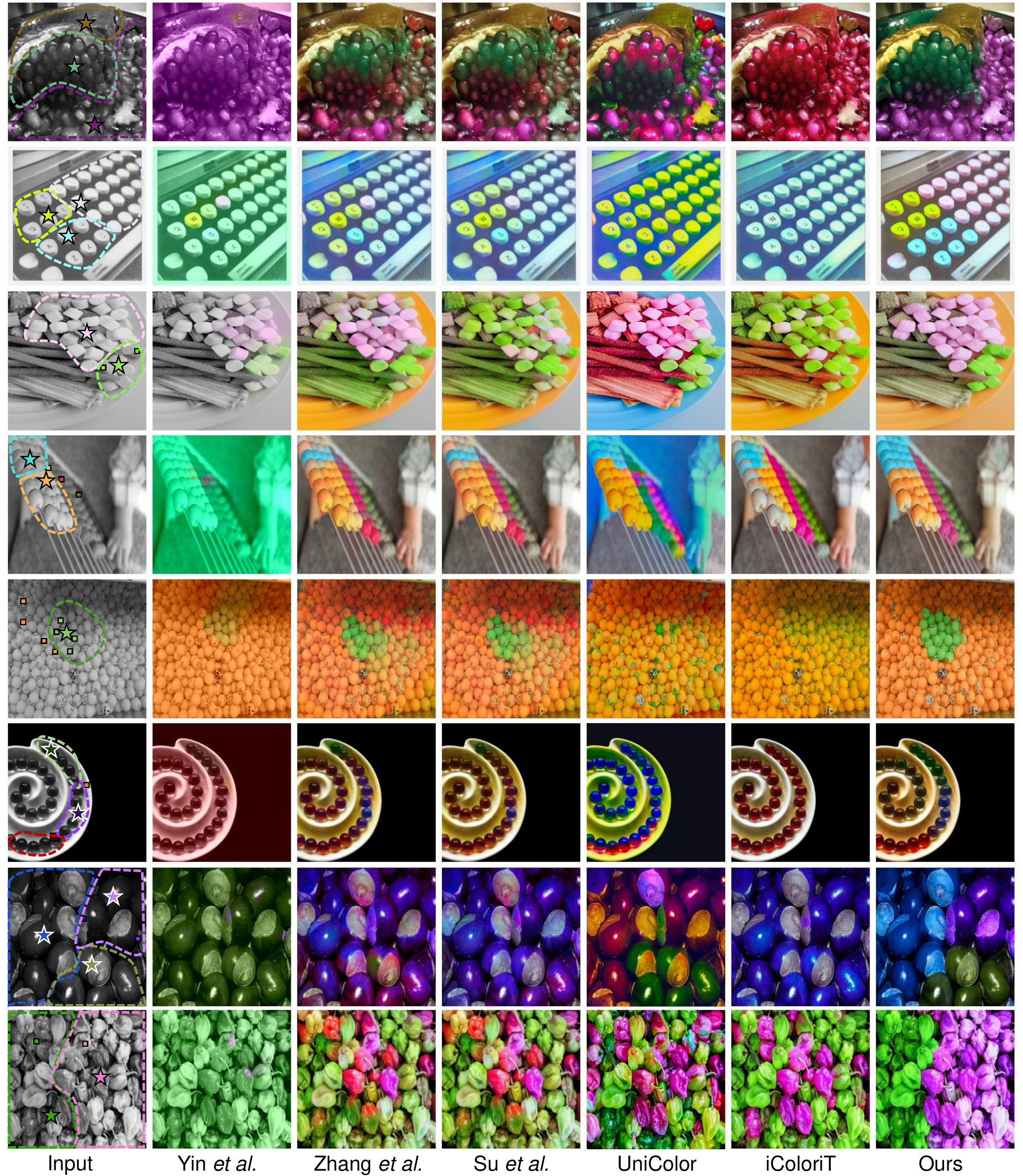}
    \caption{Qualitative results reflect the user’s preference.}
    \label{fig:supply_lasso_quali}
\end{figure*}

\noindent \textbf{Lasso with personalized colorization.} Figure~\ref{fig:supply_lasso_quali} highlights the model’s capability to generate that aligns with various user preferences. Our model is capable of producing plausible images when users wish to color similar patterns in varying colors. 

\begin{figure*}[t!]
    \centering
    \includegraphics[width=0.85\linewidth]{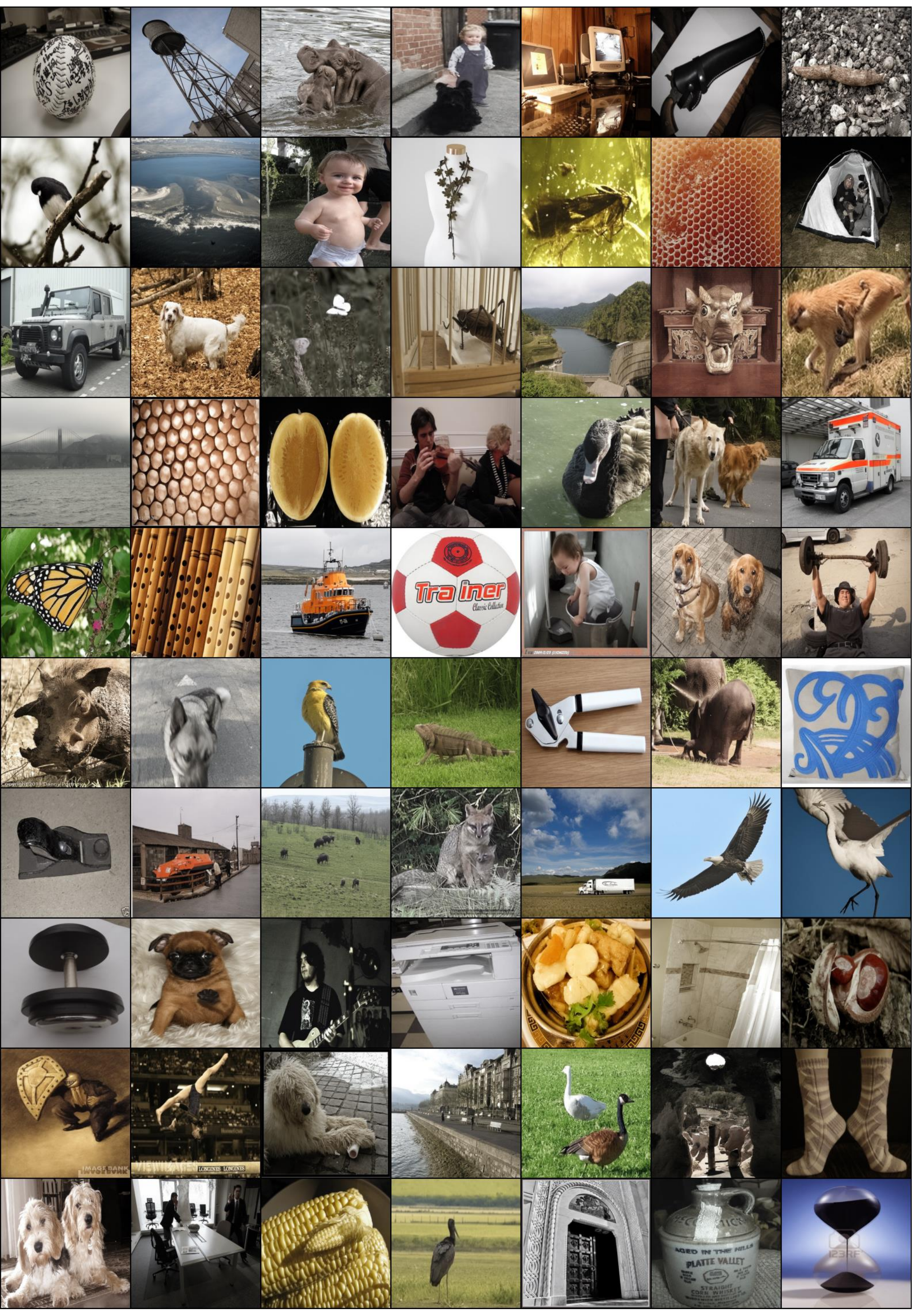}
    \caption{Uncurated colorization results on ImageNet ctest using a single hint.}
    \label{fig:supply_random_1}
\end{figure*}

\begin{figure*}[t!]
    \centering
    \includegraphics[width=0.85\linewidth]{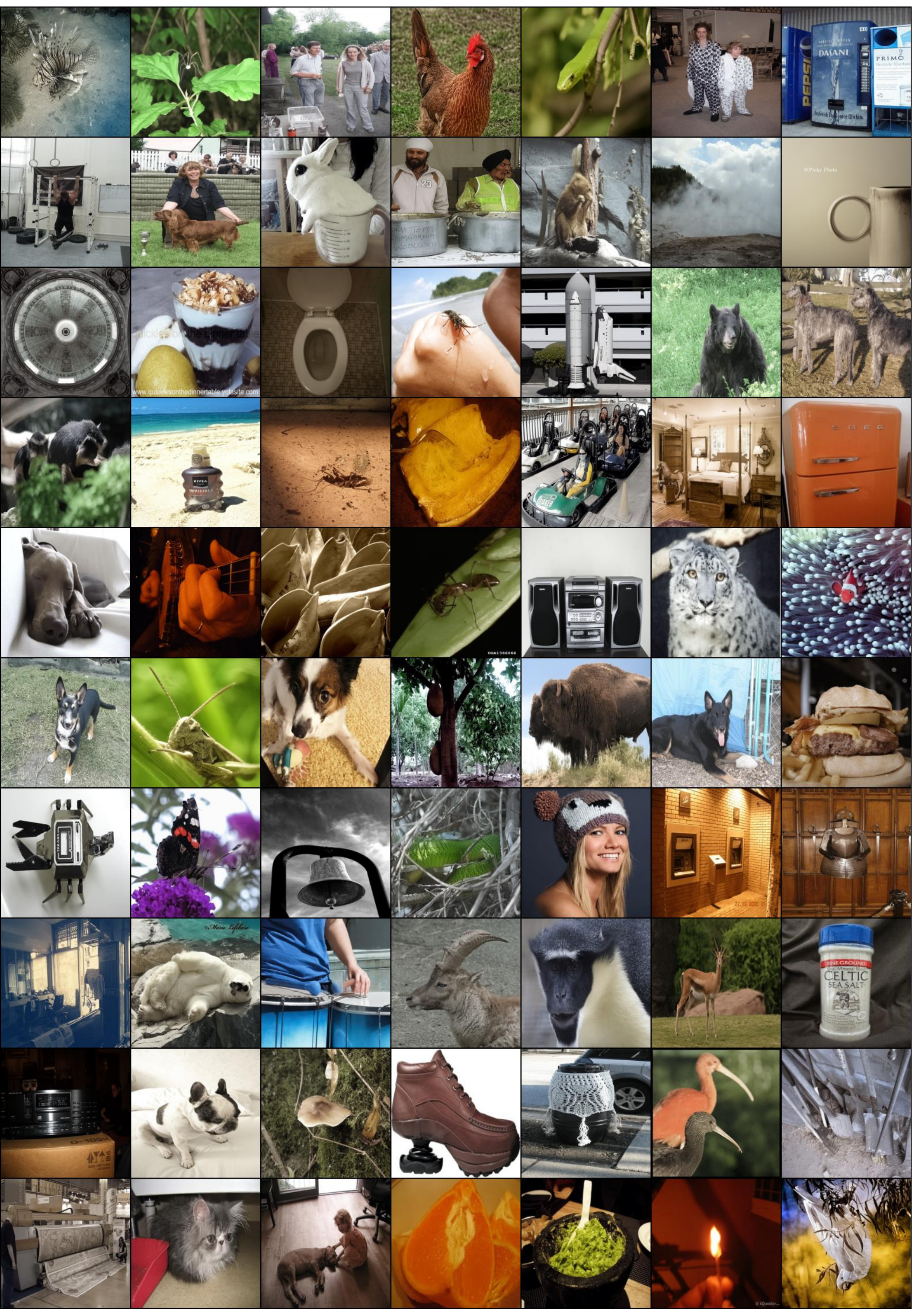}
    \caption{Uncurated colorization results on ImageNet ctest using ten hints.}
    \label{fig:supply_random_10}
\end{figure*}

\begin{figure*}[t!]
    \centering
    \includegraphics[width=0.85\linewidth]{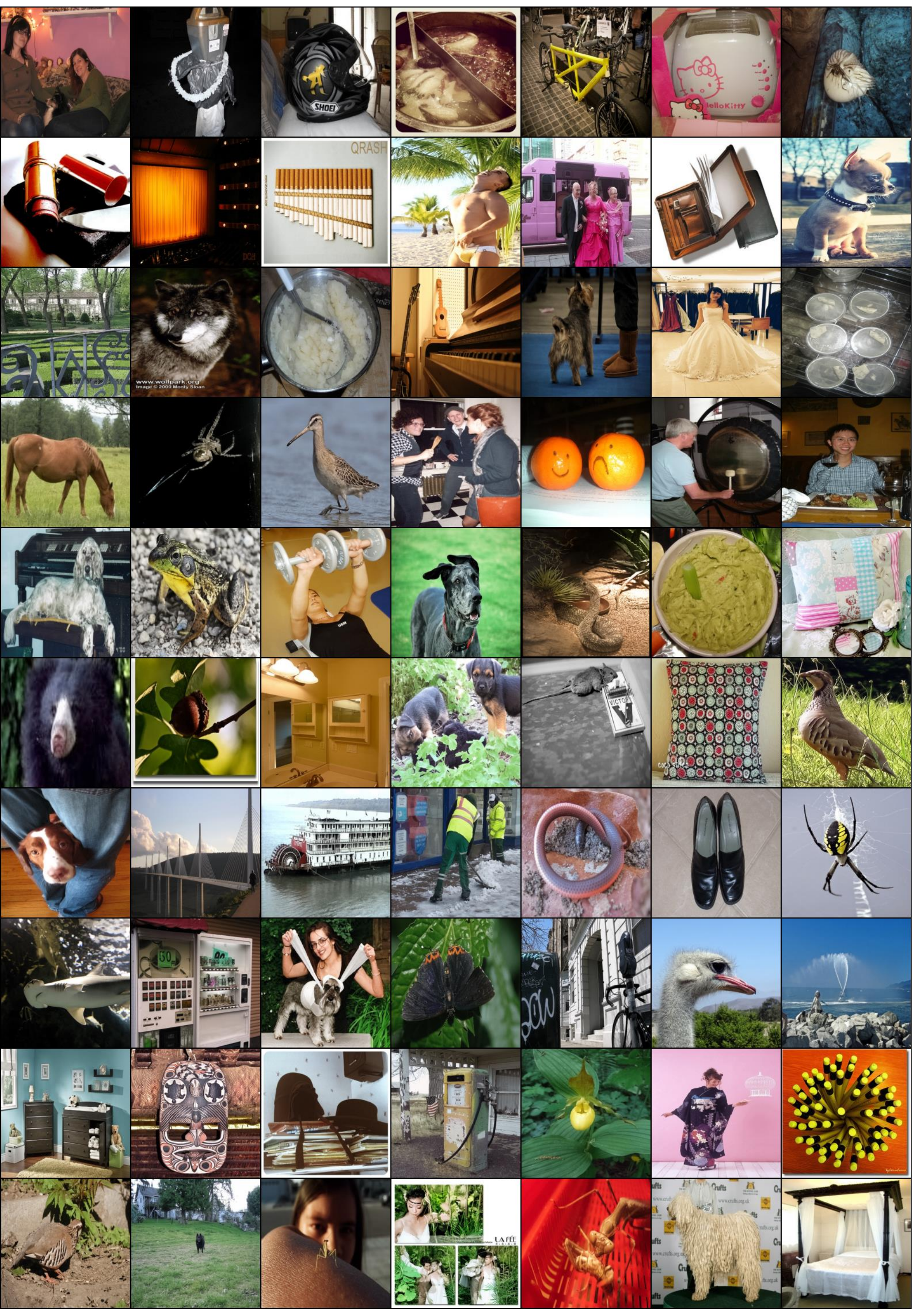}
    \caption{Uncurated colorization results on ImageNet ctest using hundred hints.}
    \label{fig:supply_random_100}
\end{figure*}

\noindent \textbf{Exploration of uncurated colorized outcomes.} Figure~\ref{fig:supply_random_1} through Figure~\ref{fig:supply_random_100} present uncurated colorized outcomes with a fixed lasso, where color hints were again sampled from the ground truth. These results illustrate that even without the challenges of color collapse, the model is capable of generating natural results.

\maketitle
\end{document}